\documentclass[journal,twoside,web]{ieeecolor}
\usepackage{jsen}
\usepackage{cite}
\usepackage{amsmath,amssymb,amsfonts}
\usepackage{algorithmic}
\usepackage{graphicx}
\usepackage{textcomp}
\usepackage{wrapfig}
\usepackage{booktabs}
\usepackage{hyperref}

\def\BibTeX{{\rm B\kern-.05em{\sc i\kern-.025em b}\kern-.08em
    T\kern-.1667em\lower.7ex\hbox{E}\kern-.125emX}}
\definecolor{abstractbg}{rgb}{0.89804,0.94510,0.83137}
\begin{document}
\title{Physical Adversarial Attack on Monocular Depth Estimation via Shape-Varying Patches}
% \author{First A. Author, \IEEEmembership{Fellow, IEEE}, Second B. Author, and Third C. Author, Jr., \IEEEmembership{Member, IEEE}
\author{Chenxing Zhao, Yang Li, \IEEEmembership{Member, IEEE}, Shihao Wu, Wenyi Tan, Shuangju Zhou, Quan Pan, \IEEEmembership{Member, IEEE}
% \thanks{This paragraph of the first footnote will contain the date on 
% which you submitted your paper for review. It will also contain support 
% information, including sponsor and financial support acknowledgment. For 
% example, ``This work was supported in part by the U.S. Department of 
% Commerce under Grant BS123456.'' }
\thanks{This work is funded by the National Natural Science Foundation of China (No.62103330, 62233014). \textit{(Corresponding author: Yang Li.)}}
% \thanks{The next few paragraphs should contain 
% the authors' current affiliations, including current address and e-mail. For 
% example, F. A. Author is with the National Institute of Standards and 
% Technology, Boulder, CO 80305 USA (e-mail: author@boulder.nist.gov). }
\thanks{The authors are with the School of Automation, Northwestern Polytechnical University, Xi’an 710129, Shaanxi, China (e-mail: liyangnpu@nwpu.edu.cn). }}

\IEEEtitleabstractindextext{%
\fcolorbox{abstractbg}{abstractbg}{%
\begin{minipage}{\textwidth}%
\begin{wrapfigure}[14]{c}{3.9in}%
\includegraphics[width=3.8in]{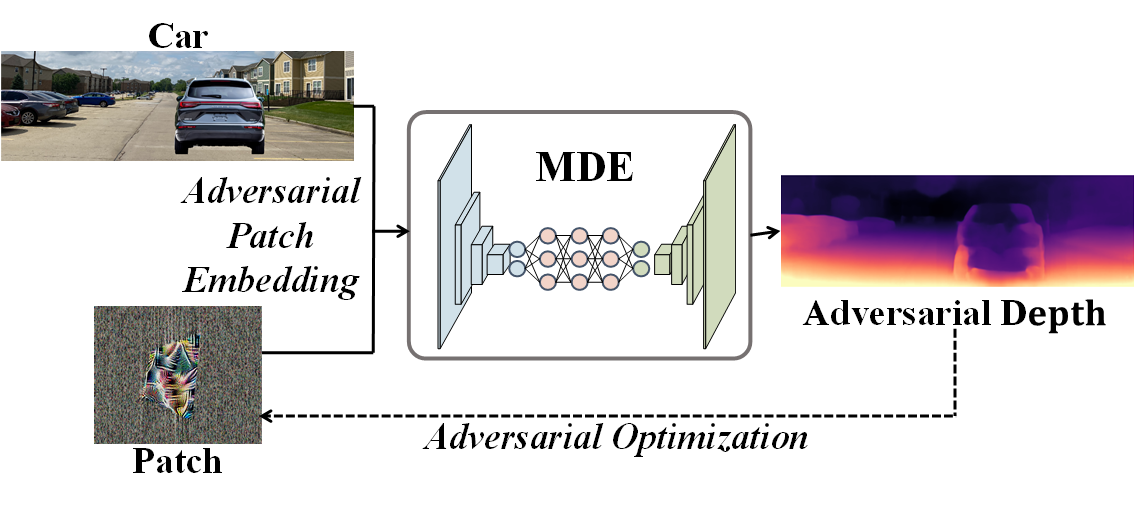}%
\end{wrapfigure}%
\begin{abstract}
Adversarial attacks against monocular depth estimation (MDE) systems pose significant challenges, particularly in safety-critical applications such as autonomous driving. Existing patch-based adversarial attacks for MDE are confined to the vicinity of the patch, making it difficult to affect the entire target. To address this limitation, we propose a physics-based adversarial attack on monocular depth estimation, employing a framework called Attack with Shape-Varying Patches (ASP), aiming to optimize patch content, shape, and position to maximize effectiveness. We introduce various mask shapes, including quadrilateral, rectangular, and circular masks, to enhance the flexibility and efficiency of the attack. Furthermore, we propose a new loss function to extend the influence of the patch beyond the overlapping regions. Experimental results demonstrate that our attack method generates an average depth error of 18 meters on the target car with a patch area of 1/9, affecting over 98\% of the target area.
\end{abstract}

\end{minipage}}}

\maketitle

\section{Introduction}
\label{sec:introduction}
\IEEEPARstart{I}{n} recent years, significant advancements in deep learning have propelled the development of the computer vision field, enabling neural networks to perceive and comprehend visual data with unprecedented accuracy and efficiency. A fundamental task in computer vision is Monocular Depth Estimation(MDE), which aims to infer the depth information of a scene from a single 2D image. In autonomous driving tasks, due to the lower cost of monocular cameras, Tesla \cite{Tesla} has integrated MDE algorithms into its autonomous driving system. Other companies, such as Huawei \cite{aich2021bidirectional}, are also actively researching this technology. Indeed, depth estimation tasks are crucial in various applications including unmanned aerial vehicles~\cite{yang2019fast, carrio2018drone}, robotics~\cite{biswas2012depth}, augmented reality~\cite{du2020depthlab}, virtual reality~\cite{el2019survey}, and scene understanding~\cite{chen2019towards,ren2019deep,armeni2017joint}.

However, despite significant progress in MDE, the vulnerability of deep learning models to adversarial attacks~\cite{DBLP:journals/corr/SzegedyZSBEGF13} remains a major concern. These attacks pose serious threats to the reliability and robustness of deep learning systems, potentially leading to critical failures in safety-critical applications.

\begin{figure*}[t!]
	\centering
	\begin{minipage}[b]{0.3\textwidth}
		\centering
		\includegraphics[width=\textwidth]{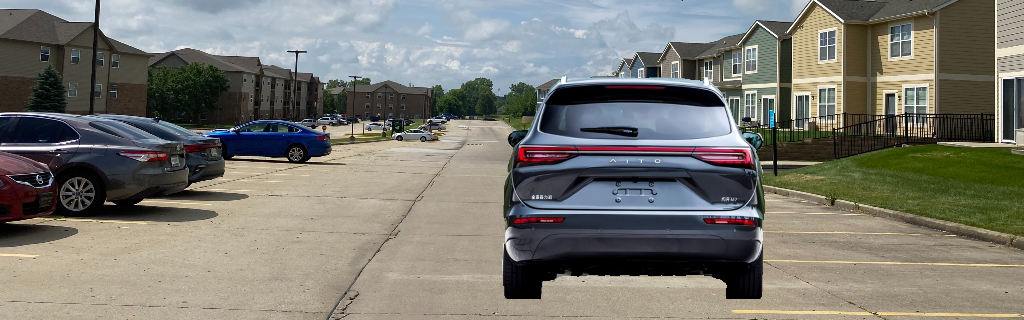}
	\end{minipage}
	\begin{minipage}[b]{0.3\textwidth}
		\centering
		\includegraphics[width=\textwidth]{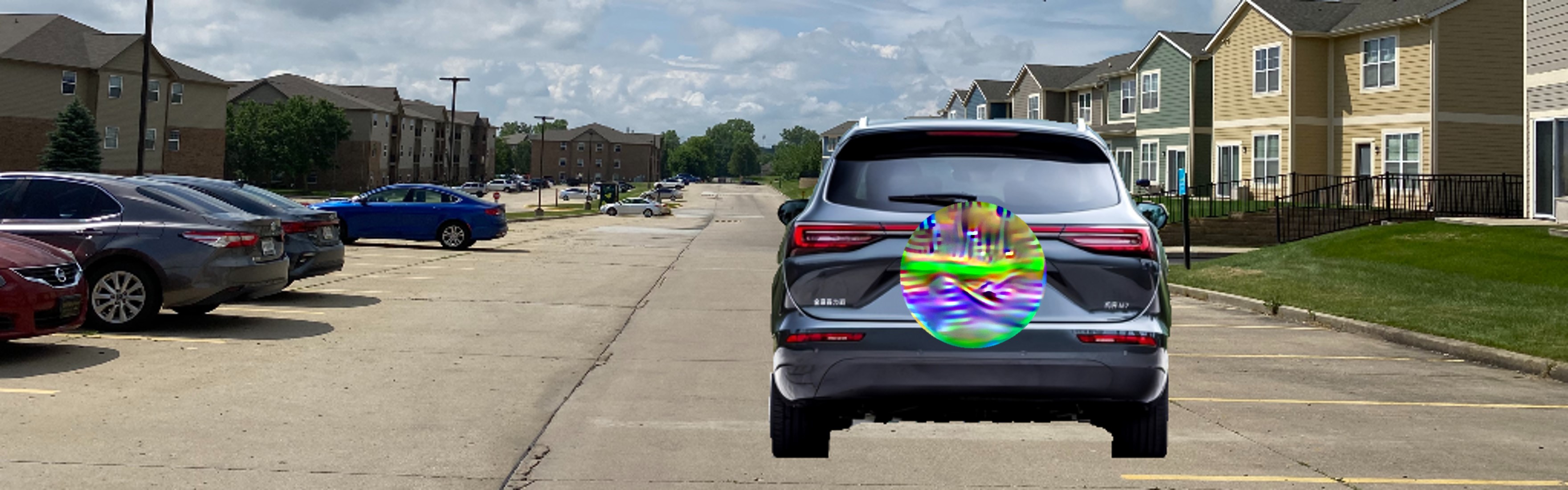}
	\end{minipage}
	\begin{minipage}[b]{0.3\textwidth}
		\centering
		\includegraphics[width=\textwidth]{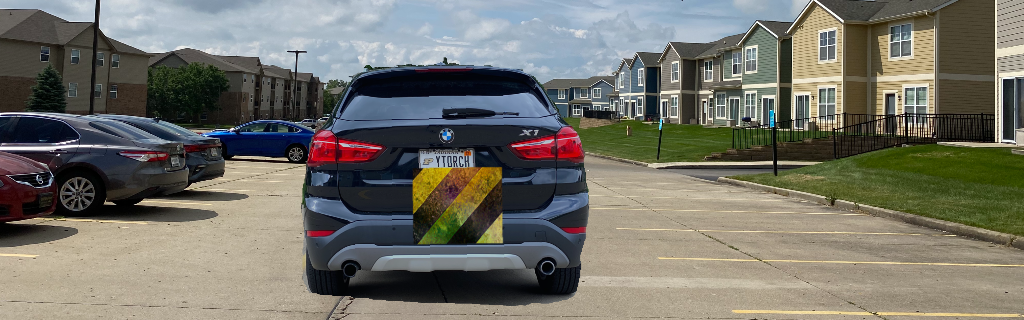}
	\end{minipage}
	\vspace{0.1cm} % 添加垂直空白
	\begin{minipage}[b]{0.3\textwidth}
		\centering
		\includegraphics[width=\textwidth]{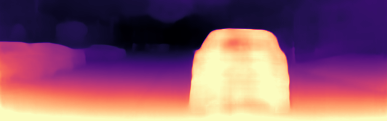}
	\end{minipage}
	\begin{minipage}[b]{0.3\textwidth}
		\centering
		\includegraphics[width=\textwidth]{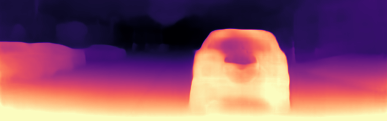}
	\end{minipage}
	\begin{minipage}[b]{0.3\textwidth}
		\centering
		\includegraphics[width=\textwidth]{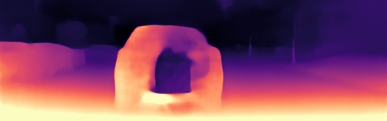}
	\end{minipage}
	\begin{minipage}[b]{0.3\textwidth}
		\centering
		(a) Benign Sample
	\end{minipage}
	\begin{minipage}[b]{0.3\textwidth}
		\centering
		(b) DisM \cite{yamanaka2020adversarial}
	\end{minipage}
	\begin{minipage}[b]{0.3\textwidth}
		\centering
		(c) StylePatch \cite{cheng2022physical}
	\end{minipage}
	\begin{minipage}[b]{0.3\textwidth}
		\centering
		\includegraphics[width=\textwidth]{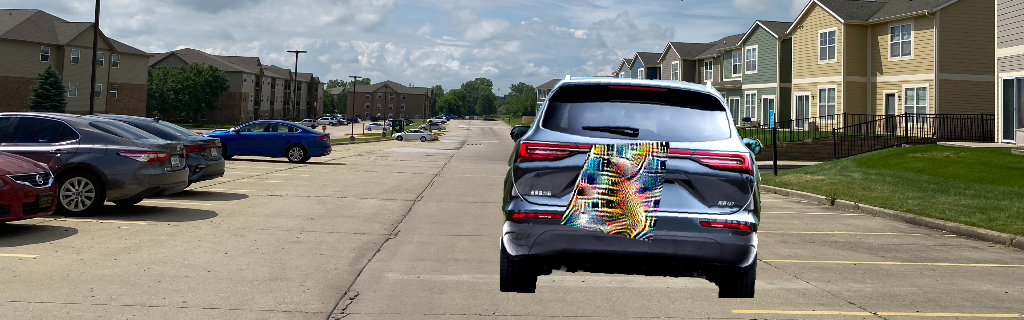}
	\end{minipage}
	\begin{minipage}[b]{0.3\textwidth}
		\centering
		\includegraphics[width=\textwidth]{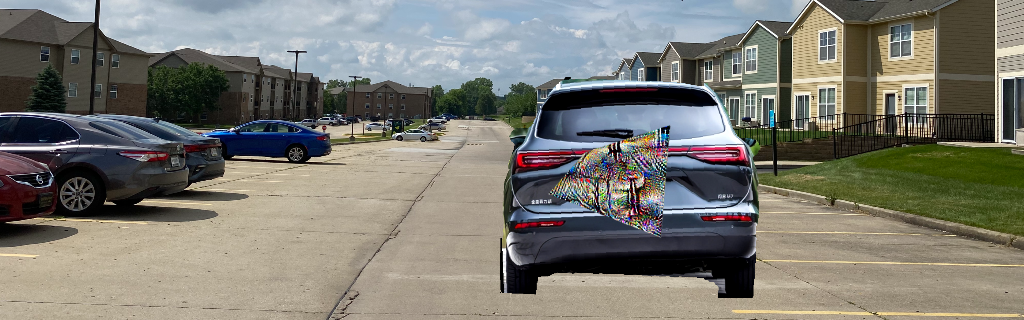}
	\end{minipage}
	\begin{minipage}[b]{0.3\textwidth}
		\centering
		\includegraphics[width=\textwidth]{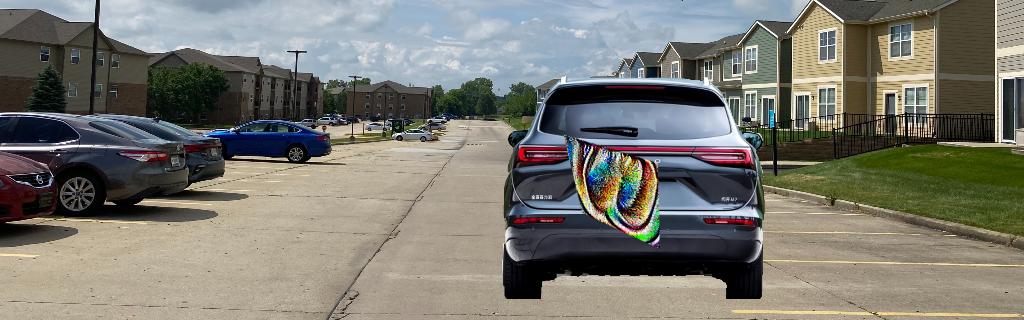}
	\end{minipage}
	\vspace{0.1cm} % 添加垂直空白
	\begin{minipage}[b]{0.3\textwidth}
		\centering
		\includegraphics[width=\textwidth]{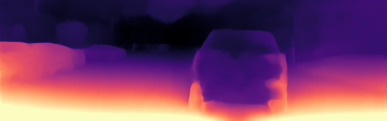}
	\end{minipage}
	\begin{minipage}[b]{0.3\textwidth}
		\centering
		\includegraphics[width=\textwidth]{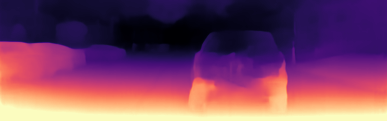}
	\end{minipage}
	\begin{minipage}[b]{0.3\textwidth}
		\centering
		\includegraphics[width=\textwidth]{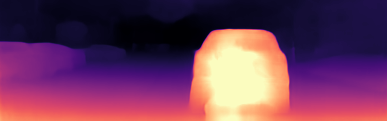}
	\end{minipage}
	\begin{minipage}[b]{0.3\textwidth}
		\centering
		(d) Further
	\end{minipage}
	\begin{minipage}[b]{0.3\textwidth}
		\centering
		(e) Disappear
	\end{minipage}
	\begin{minipage}[b]{0.3\textwidth}
		\centering
		(f) Closer
	\end{minipage}
	\caption{The image in (a) depicts benign content, wherein the mean depth of the car is measured at 6.77 meters; The patch shown in (b) was directly obtained from the source code provided in Distribution Model (DisM for short)~\cite{yamanaka2020adversarial}. This patch introduces a mean depth error of 1.45 meters to the car; The patch displayed in (c) was trained using the source code provided in StylePatch Model~\cite{cheng2022physical}. This patch results in a mean depth error of 7.19 meters affecting the car; (d) is the patch trained to maximize the distance of the car, and the mean depth error reaches 19.17 meters; In (e), our objective was to make the car appear to vanish within the depth map, rather than the depth map simply missing a portion of its content; In (f), we aimed to bring the car closer. As a result, the mean depth of the car is reduced to 5.05 meters.}
	\label{fig:one}
\end{figure*}

Existing attacks against MDE can be classified into digital-space attacks \cite{wong2020targeted,mopuri2018generalizable} and physical-world attacks \cite{guesmi2023aparate,cheng2022physical,yamanaka2020adversarial}. Digital-space attacks primarily target deep learning models in computer environments, where attackers manipulate input images with subtle modifications to deceive the model, resulting in incorrect depth estimation outputs. In contrast, physical-world attacks occur in real-world environments, where attackers disturb the input of depth estimation models by adding special textures, stickers, or other physical devices to input images, leading to erroneous depth estimation results.

Physical-world attacks pose more challenges and significance compared to digital-space attacks because they need to consider more real-world constraints and factors~\cite{athalye2018synthesizing}. Factors such as lighting conditions, surface reflectance properties of objects, camera positions, and angles can influence the performance of depth estimation models in real scenes, requiring attackers to consider these factors when designing effective attack methods comprehensively. Although physical-world attacks are more challenging, they also have a more direct and severe impact on the performance and reliability of depth estimation models. In practical applications, incorrect depth estimation results may lead to serious safety accidents or economic losses, underscoring the importance of research on physical-world attacks.

Apart from the camouflage-based adversarial attack~\cite{li2024flexible}, the patch-based adversarial attack is another usually used approach to conduct the physical-world attack~\cite{tan2023doepatch}. However, the existing patch-based adversarial attack methods exhibit limited scope, primarily confining the influence of the patch within its designated region. Furthermore, the prevalent utilization of a single rectangular or circular patch in current adversarial patch works overlooks the potential impact of patches with varying shapes.
% Our objective is to expand the coverage of patch-based attacks across the entire target object while refining the patch's position and shape to achieve optimal effectiveness. To this end, we propose a comprehensive framework aimed at enhancing the efficacy of patch attacks by optimizing the content, shape, and position of patches directed at specific objects. This multifaceted framework provides various attack modes, facilitating the generation of inaccurate depth information or complete concealment of the target object.
In response to these limitations, we have proposed a new adversarial loss function and explored the impact of different patch shapes on the effectiveness of the attack. Our objective is to expand the coverage of patch-based attacks to encompass the entire target object, while refining the patch's position and shape to achieve optimal effectiveness. To this end, we propose a comprehensive framework aimed at enhancing the efficacy of patch attacks by optimizing the adversarial texture, shape, and position of patches tailored to specific objects. This multifaceted framework offers various attack modes aimed at generating inaccurate depth information or completely concealing the target object. In summary, the contributions of this work are as follows:

\begin{itemize}
	\item We have designed a novel patch-based depth attack framework that effectively controls the attack distance while expanding the attack's impact range.
	% We have devised a novel attack framework capable of inducing various incorrect perceptions of depth regarding the target object, thereby causing depth of the target car to appear closer, farther away, or even disappearing into the background.
	\item We have developed a new method for generating adversarial patches with different shapes and validated their effectiveness on depth estimation adversarial attacks.
	% We have developed a novel method for generating patch masks and investigated the influence of patch shapes on the effectiveness of the attack.
	% \item We have designed a new loss function which is to extend the influence region of the patch to encompass the entire target object.
	\item Our experiments, conducted from digital-space to physical-world, consistently confirm the effectiveness of our proposed method. The introduced patch yields an average depth error exceeding 7 meters.
	% The proposed patch achieves a mean depth error exceeding 18 meters and affects over 98\% of the target area.
\end{itemize}

\section{Related Work}
\subsection{Sensor Attacks on Autonomous Driving}

In the field of autonomous driving, previous adversarial works have mainly focused on directly interfering with the detection of hardware sensors. For example, Yan et al.~\cite{yan2016can} deployed contactless attacks on millimeter-wave radar, and Xu et al.~\cite{xu2018analyzing} developed random spoofing, adaptive spoofing, and jamming attacks on ultrasonic sensors. Petit et al.~\cite{Petit2015RemoteAO} investigated attacks on sensors such as cameras and LiDAR from external sources in the automotive context. Shin et al.~\cite{shin2017illusion} introduced a spoofing by relaying attack, inducing illusions in lidar output and making them appear closer than the location of the spoofing device. Whereas Cao et al.~\cite{cao2019adversarial} involved placing objects in the physical-world to disrupt LiDAR detection. Furthermore, numerous studies focus on attacking GPS~\cite{papadimitratos2008protection, humphreys2008assessing, qiu2020semanticadv}. Recently, there has been a shift towards interfering with software algorithm~\cite{athalye2018synthesizing, sitawarin2018darts}, which involves adding human-imperceptible noise to images in digital-space or placing adversarial objects in the physical-world in front of the camera to introduce errors in detection and recognition models.

% which involves placing adversarial objects in front of cameras to induce errors in detection and recognition models. 

In these algorithmic adversarial attacks, particular emphasis is placed on adversarial attacks targeting MDE models. Adversarial attacks in the digital-space against MDE can induce significant changes in the depth of the entire image through noise interference, whereas the depth range of the impact of physical domain patch attacks still remains around the patch.

\begin{figure*}[t]
	\centering
	\includegraphics[width=\textwidth]{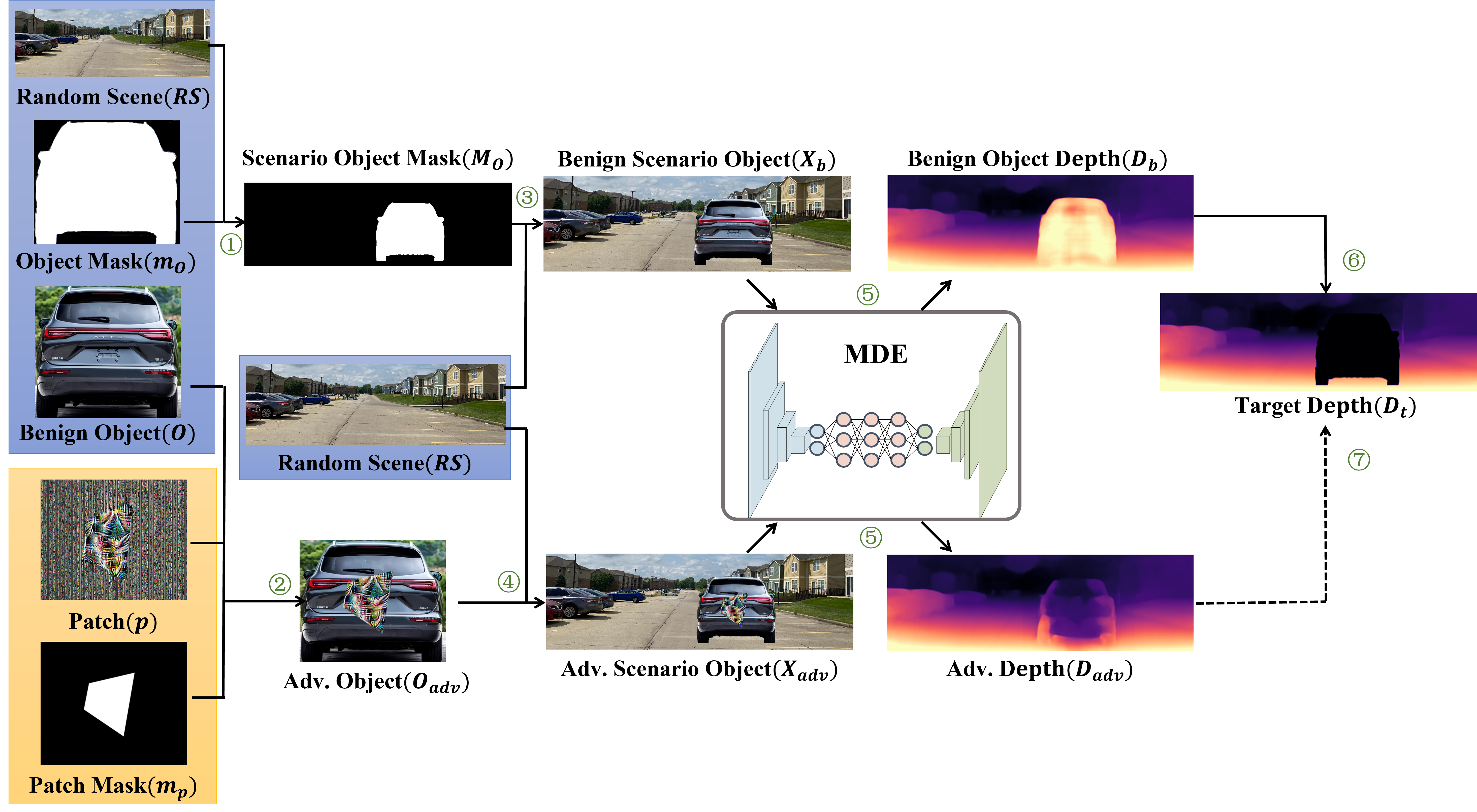}
	\caption{Overview of the framework. The inputs is in the blue box, which includes the Benign Object ($O$), Object Mask ($m_O$), and Random Scene ($RS$). The Patch ($p$) and Patch Mask ($m_p$) in the yellow box represent the content we aim to optimize. Please refer to section~\ref{sec:overview} for detailed information regarding the framework diagram.}
	\label{fig:overview}
\end{figure*}
\subsection{Depth Estimation Adversarial Attacks}

In the field of computer vision, adversarial attacks have predominantly concentrated on physical attacks aimed at object detection~\cite{thys2019fooling, hu2021naturalistic, tan2023doepatch, li2024flexible}, image classification~\cite{athalye2018synthesizing, song2018physical, li2023few}, face recognition~\cite{sharif2016accessorize, komkov2021advhat}, and similar tasks~\cite{LI2024future}. However, research efforts on adversarial attacks against depth estimation, comparatively, have been relatively sparse.

Adversarial attacks on MDE can be specifically categorized into two major types: digital-space attacks and physical-world attacks.
In the digital-space attack, it is usually to generate the adversarial noise perturbation. 
For example, Dai et al.~\cite{daimo2021black} treated MDE models as black boxes and employed a multi-objective optimization approach for the perturbation generation. Mopuri et al.~\cite{mopuri2018generalizable} proposed a multi-task attack method by learning global noise perturbations. 
Later, Wong et al.~\cite{wong2020targeted} and Zhang et al.~\cite{zhang2020adversarial} adapted classical classification attack methods to target depth estimation.
% Wong et al.~\cite{wong2020targeted} had improved classical classification attack methods for the purpose of attacking depth estimation, while Zhang et al.~\cite{zhang2020adversarial} improved classical classification attack methods for attacking depth estimation. 

In contrast, physical-world adversarial attacks are more intricate due to real-world constraints. 
For example, Guesmi et al.~\cite{guesmi2024saam} focused on indoor scenes, augmenting common images with adversarial noise and printing them to hang indoors to disrupt household robot depth estimation. 
Yamanaka et al.~\cite{yamanaka2020adversarial} trained depth-fixed patches, maintaining constant depth regardless of environmental variations, albeit without optimization for specific target cars.
% Yamanaka et al.~\cite{yamanaka2020adversarial} trained depth-fixed patches, maintaining constant depth regardless of environmental depth variations. However, the effectiveness of these patches is limited to the patch area and lacks optimization for specific target cars, being optimized for random positions instead. 
Cheng et al.~\cite{cheng2022physical} conducted style transfer on patches and optimized their positions and shapes, though the effects are confined to the patch area and rarely cover the target car position. Moreover, optimization is limited to rectangular shapes. Similarly, the patches proposed by Guesmi et al.~\cite{guesmi2023aparate} cover a larger area but lack optimization for the shape of the target car.
Different from previous works, our objective is to ensure that the effects of the patch cover the entire target car while only influencing its depth, without causing disturbance to the surrounding depths.

\section{Proposed Method}

\subsection{Problem Formulation}

This section mainly describes the problem of adversarial attacks on depth estimation using patches. The problem can be described by the following equation:
% \begin{equation}
	% \min_{S(P)}\left\|P\right\|_ps.t.F(O_{adv})\neq F(O)
	% \end{equation}
\begin{equation}
	\min_{p} S(p) \quad  \text{s.t.} \quad  F(O_{adv}) \neq F(O)
\end{equation}
where $p$ represents the patch, $S(p)$ represents the area of $p$, $F$ represents the MDE network, $O$ represents the original benign sample, and $O_{adv}$ represents the generated adversarial image, which can be expressed as follows: 
\begin{equation}
	O_{adv}=(1-m_p)\odot O+m_p\odot p 
	\label{con:1}
\end{equation}
where $m_p$ represents the patch mask, $\odot$ denotes element-wise multiplication. Since the MDE model is non-convex, a closed-form solution to this problem cannot be found. Therefore, we reformulate the problem as follows:
\begin{equation}
\arg\max_p\sum_{O\in D}L\left(F(O_{adv}),F(O)\right)
\end{equation}
where $L$ is the mean square error loss function, and $D$ represents the training dataset. We term the method aiming to push the adversarial deep graph as far away as possible from the benign deep graph as non-targeted attack. After the transformation, the optimal solution $p$ can be found by reducing the loss function $L$ and optimizing the pixels of the patch $p$. Certainly, in addition to the optimization approach mentioned above, we can also introduce a target image \(O_{target}\) and optimize \(O_{adv}\) to approximate \(O_{target}\):
\begin{equation}
\arg\min_p\sum_{O\in D}L\left(F(O_{adv}),F(O_{target})\right)
\end{equation}
We term the method providing a specific target deep graph as targeted attack, and targeted attack method is uniformly adopted in this paper.

\begin{figure}[tbp]
	\centering
	\begin{minipage}[b]{0.28\textwidth}
		\centering
		\includegraphics[width=\textwidth]{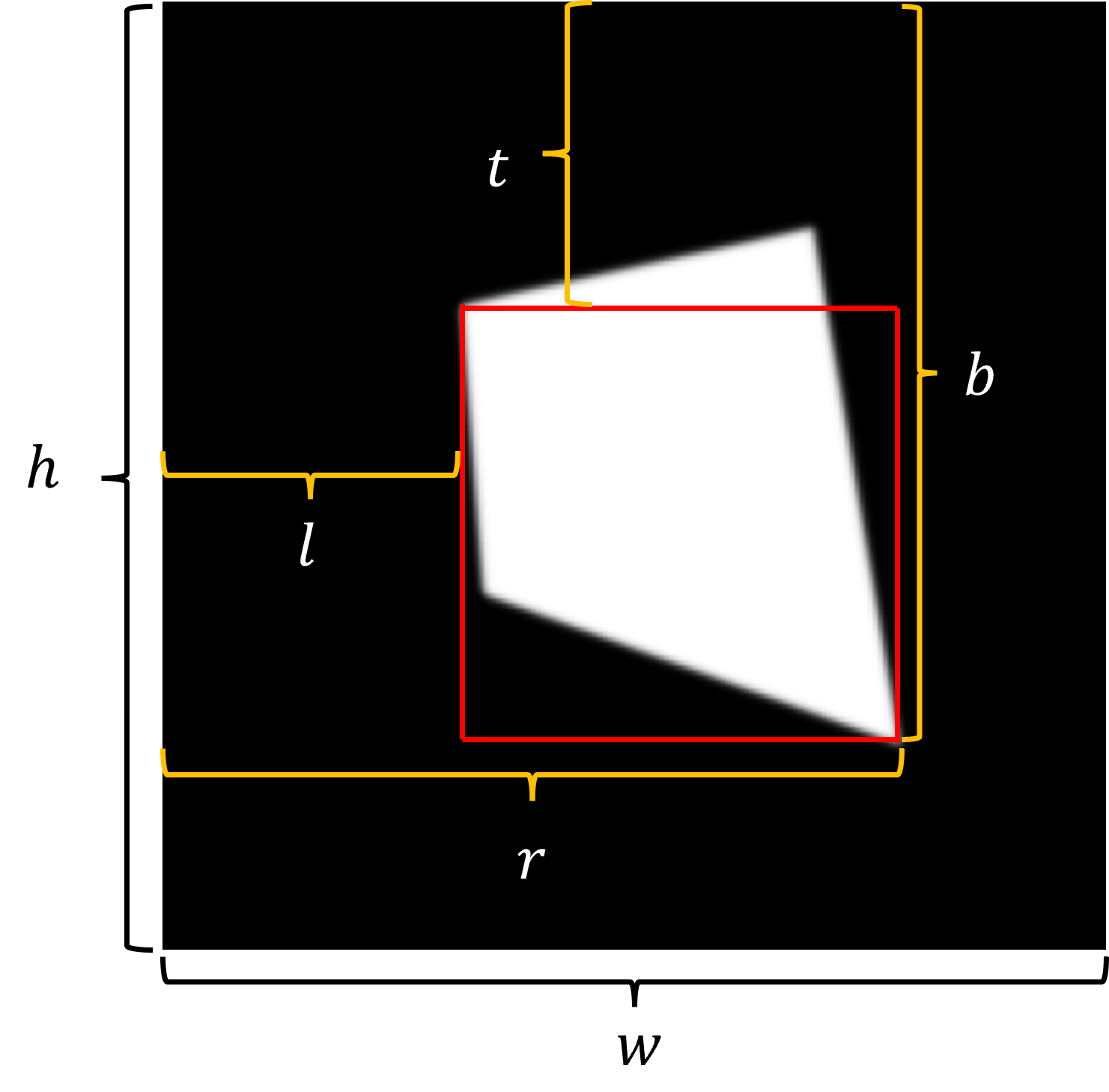}
	\end{minipage}
	\caption{Arbitrary Quadrilateral Mask. Wherein, \(h\) and \(w\) respectively represent the height and width of the mask, and \({\Theta_1} = [l, r, t, b]\) are the boundary parameters we set.}
	\label{fig:quadrilateral}
\end{figure}

\subsection{ASP Framework}
\label{sec:overview}

Our model framework is illustrated in Fig.\ref{fig:overview}. The patch \( p \) and patch mask \( m_{p} \) in the yellow box represent the content we aim to optimize. The input consists of the target car image \( O \) the target car mask \( m_c \), and random scene \( RS \) in the blue box. To enhance the diversity of the dataset and the adaptability of the models, we adopt the method proposed in Cheng et al.~\cite{cheng2022physical}, which involves embedding car images into various background images. In Fig. \ref{fig:overview}, as illustrated in \textcircled{1}, we apply random transformations including size, position, rotation, and brightness to the car, please refer to section~\ref{data_pre} for details. The car's mask is resized to match the size of the random scene \( RS \) in the dataset, forming the scenario car mask \(M_O\). As depicted in \textcircled{2}, we apply the transformation shown in \eqref{con:1} to the patch \( p \) , car patch mask \( m_p \) and benign target \( O\) to generate adversarial object \( O_{adv} \). Following this, in \textcircled{3}, the benign scenario image \( X_{b} \) is constructed by embedding the benign target \( O \) into the random scene \( RS \) via the mask \( M_{O} \). We then proceed to \textcircled{4}. The adversarial scenario image \( X_{adv} \) is formed by pasting the patch target car onto the random scene \( RS \). Moving on to \textcircled{5}, \( X_{adv} \) and \( X_{b} \) are input to the MDE model to obtain \( D_{adv} \) and \(D_b\), where \( D_{adv} \) and \(D_b\) represent the depth maps of adversarial samples and clean samples respectively. We then undertake \textcircled{6}, where based on different attack modes, we set the depth of the targeted car portion to various depth maps \(D_t\), please refer to section~\ref{sec:attack mode} attack mode for details. Finally, in \textcircled{7}, we optimize \(D_{adv}\) based on \(D_t\) using adversarial loss function. The loss function will be detailed in section~\ref{loss}.

\subsection{Optimization of Patch Shape and Position}

In prior literature, scant attention has been devoted to examining the impact of mask shape and position on the efficacy of attacks. To address this gap, we have proposed optimization methods aimed at exploring diverse mask shapes and positions. Specifically, we have formulated strategies for generating rectangular, quadrilateral, and circular masks. 

\subsubsection{Quadrilateral}

\begin{figure}[tbp]
	\centering
	\begin{minipage}[b]{0.15\textwidth}
		\centering
		\includegraphics[width=\textwidth]{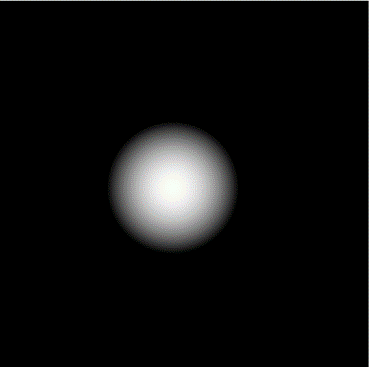}
	\end{minipage}
	\begin{minipage}[b]{0.15\textwidth}
		\centering
		\includegraphics[width=\textwidth]{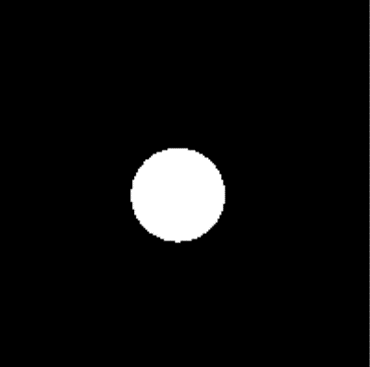}
	\end{minipage}
	\begin{minipage}[b]{0.15\textwidth}
		\centering
		\includegraphics[width=\textwidth]{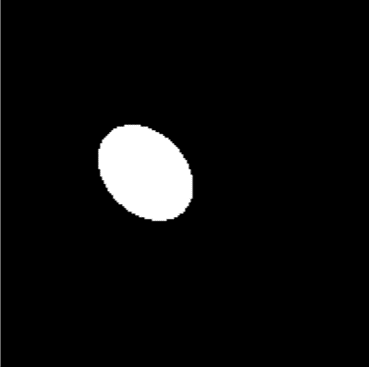}
	\end{minipage}
	\begin{minipage}[b]{0.15\textwidth}
		\centering
		(a)
	\end{minipage}
	\begin{minipage}[b]{0.15\textwidth}
		\centering
		(b)
	\end{minipage}
	\begin{minipage}[b]{0.15\textwidth}
		\centering
		(c)
	\end{minipage}
	\caption{Circle Mask. (a) Circular Mask without Binary; (b) Circular Mask with Binary; (c) Oval Mask}
	\label{fig:circle}
\end{figure}

Since patch masks are typically binary 0-1 masks, they are non-differentiable and cannot be directly optimized. To address this issue, Cheng et al.~\cite{cheng2022physical} proposed a mask optimization framework that enables patches to produce gradients and be optimized accordingly. On a mask with width \( w \) and height \( h \), they define four boundary parameters \( {\Theta_1} = [l, r, t, b] \), and \( 0 \leq l \leq r \leq w \) and \( 0 \leq b \leq t \leq h \), see Fig.\ref{fig:quadrilateral}, \( l \) and \( r \) represent the boundary values of the left and right edges of the rectangle, while \( t \) and \( b \) represent the boundary values of the top and bottom edges of the rectangle. The formula for calculating the patch mask for a rectangle is as follows, where \( i \) and \( j \) represent the pixel coordinates in the grid and \(i \in [1...w],\, j \in [1...h]\):

\begin{equation}
	\begin{aligned}\label{con:2}
		m_p[i,j] = & \frac{1}{4} \left[ - \tanh (i - t) \cdot \tanh (i - b) + 1 \right] \\
		& \cdot \left[ - \tanh (j - l) \cdot \tanh (j - r) + 1 \right] 
	\end{aligned}
\end{equation}
However, the aforementioned formula can only generate a rectangular mask with a fixed shape. We have proposed a new patch generation method capable of producing arbitrary quadrilateral patch shapes. We define four slope parameters for the quadrilateral \( S = [s_{l}, s_{r}, s_{t}, s_{b}] \), and slope parameters \( S \) are within the range [-1, 1], our formula is as follows:
	\begin{flalign}
 \begin{aligned}
		 m_p[i,j] & = \frac{1}{4} \left\{ - \tanh \left[ i - \left( t + s_t \cdot \left( j - l \right) \right) \right] \cdot \right.\\
		& \quad \left. \tanh \left[ i - \left( b + s_b \cdot \left( j - r \right) \right) \right] + 1 \right\} \\
		& \quad \cdot \left\{ - \tanh \left[ j - \left( l + s_l \cdot \left( i - t \right) \right) \right] \cdot \right.\\
		& \quad \left. \tanh \left[ j - \left( r + s_r \cdot \left( i - b \right) \right) \right] + 1 \right\} 
 \end{aligned}
	\end{flalign}
where, taking \( t + s_t(j - l) \) as an example, it is regarded as a whole, contrasted with \( t \) in \eqref{con:2}, except when the pixel position is at \((t,l)\), at all other pixels, a linear change will occur, effectively causing the value of \( t \) in the mask to linearly vary with the movement of the pixel, thus forming a skewed boundary. By introducing the distance from the \( l \) boundary, the angles of the corners of the mask are kept within a reasonable range, which is conducive to optimization. We treat the four boundary parameters \({\Theta_1}=[l,r,t,b]\) and the four slope parameters \(S=[s_l,s_r,s_t,s_b]\) as optimization parameters.

\begin{table*}[tbp]\centering
	\caption{Comparison of Model Attack Effects. The APARATE, StylePatch, and 
 DisM are predecessors' models. $L1$ and $L2$ denote the results obtained by employing solely one component of $L_{depth}$. InitRt represents the rectangular mask after initialization, without undergoing positional and shape optimization. DE and Gaussian are two baselines proposed in our study.}
	\label{tab:attack model result}
	\begin{tabular}{cccccc}
		\toprule
		Attack Model&$MSE_t$&$MSE_b$&$\alpha$&$\varepsilon_{disp}$&$\varepsilon_{depth}$\\
		\midrule
		APARATE\cite{guesmi2023aparate} & 39.68 & 147.14 & 96.41\% & 0.6548 & 16.68\\
		StylePatch\cite{cheng2022physical}& 47.51 & 112.43 & 82.68\% & 0.3973 & 7.19\\
		DisM\cite{yamanaka2020adversarial}& 250.25 & 6.81 & 44.28\% & 0.1467 & 1.45\\
		$L_1$ & 93.81 & 72.02 & 95.37\% & 0.4536 & 7.34\\
		$L_2$ & 33.89 & 152.47 & 97.54\% & 0.6675 & 17.72\\
		InitRt & 32.24 & 162.61 & 97.54\% & 0.6882 & 16.53\\
		DE & 161.30 & 10.16 & 46.09\% & 0.2191 & 2.93\\
		Gaussian & 63.33 & 84.42 & 91.30\% & 0.5283 & 14.31\\
		Ours & \textbf{30.30} & \textbf{167.07} & \textbf{98.16\%} & \textbf{0.6969} & \textbf{19.18}\\
		\bottomrule
	\end{tabular}
\end{table*}

\subsubsection{Circle}

The difference between circular masks and rectangular masks lies in the fact that the former only requires three parameters: the coordinates of the center ${\Theta}_2=(x,y)$ and the radius $R$. When generating the mask, values are assigned based on the distance of each pixel to the center of the circle. The formula is as follows:
\begin{equation}
	\begin{aligned}
		& m_p[i,j] = F_B \left\{ \max \left[ 1 - \frac{1}{2} \frac{{(i - x)^2 + (j - y)^2}}{{R^2}} \right], 0 \right\} 
	\end{aligned}
\end{equation}
where $F_B$ is a binarization method specifically designed for gradient backpropagation. During the forward pass, $F_B$ performs a non-differentiable binarization operation. Specifically, it rounds each element in the input tensor to the nearest integer value. During the backward pass, the backward method simply passes gradients to downstream nodes without any modifications. This means that for this method, gradient propagation remains unchanged, i.e., the input gradient equals the output gradient. As illustrated in Fig.\ref{fig:circle}, (a) represents the non-binarized circular mask, while (b) represents the circular mask version after binarization.

Indeed, should there be a desire to modify the formula to create ellipses, we can achieve this by introducing an offset to the center coordinates based on the positions of the pixels, thereby transforming the circular mask into an elliptical mask. For instance, we may amend the formula as follows:
\begin{equation}
	\begin{aligned}
		 m_p[i,j] = & F_B \left\{ \max \left[ 1 - \frac{1}{2} \frac{{\left(i - \left(x + s_x\left(j - y\right)\right)\right)^2}}{{R^2}} \right. \right. \\
		& \left. \left. + \frac{{(j - (y + s_y(i - x)))^2}}{{R^2}} \, , \, 0 \right] \right\} 
	\end{aligned}
 \label{equ:fb}
\end{equation}
In contrast to the conventional formula for generating standard oval masks, as depicted below, our method addresses the limitation where the focus of the oval consistently resides on the coordinate axes, thereby resulting in more variability in the shape and position of the mask.
\begin{equation}
	\begin{aligned}
		& m_p[i,j] = F_B \left\{ \max \left[ 1 - \frac{1}{2} \left( \frac{{(i - x)^2}}{{a^2}} + \frac{{(j - y)^2}}{{b^2}} \right) \, , \, 0 \right] \right\} 
	\end{aligned}
\end{equation}

\begin{table*}[tbp]
    \centering
	\caption{The effects of different shaped masks under two target depths. In the table, ``Mode'' represents the attack mode, where two attack modes are considered: the ``Further'' attack mode, which aims to maximize the distance of the car from the observer, and the ``Disappear'' attack mode, which aims to make the car's depth blend into the scene's depth. The symbols ``Q'', ``Rt'', and ``C'' denote quadrilateral, rectangular, and circular masks, respectively. }
	\label{tab:at}
	\begin{tabular}{ccccccc}
		\toprule
		Mode & Shape & $MSE_t$ & $MSE_b$ & $\alpha$ &$\varepsilon_{disp}$&$\varepsilon_{depth}$\\
		\midrule
		& Q & 30.25 & 167.39 & \textbf{98.53\%} & 0.6939 & 19.18 \\
		Further	& Rt & 30.30 & 167.07 & 98.16\% & \textbf{0.6969} & 19.18 \\
		& C & \textbf{30.01} & \textbf{167.92} & 98.51\% & 0.6957 & \textbf{19.27} \\
		\midrule
		& Q & \textbf{11.72} & \textbf{98.85} & 94.87\% & \textbf{0.5287} & 11.66 \\
		Disappear & Rt & 14.43 & 91.15 & \textbf{96.32\%} & 0.5186 & 9.54 \\
		& C & 12.93 & 91.38 & 94.37\% & 0.5226 & \textbf{11.67} \\
		\bottomrule
	\end{tabular}
\end{table*}

\subsection{Adversarial Loss}
\label{loss}

We have noticed that in previous studies, although the depth in the patch region was effectively modified, areas outside the patch region were hardly affected. To address this issue, we have proposed a new loss function. Through our experiments, we observed that the depth within the overlapping regions of the patch has a significant impact, while the regions outside the patch change slowly. We aim to synchronize the rate of depth variation in the regions outside the patch with that of the overlapping patch regions. This synchronization would enable the depths outside the patch to be optimized concurrently with those of the patch. Our loss function consists of two parts: the loss function for the mask region and the loss function for the target car region outside the mask. The loss function for the mask region is defined as:
\begin{equation}L_1=\left[\left(D_{adv}-D_t\right)\odot M_p\right]^2\end{equation}
where $M_p$ represents the adversarial scenario patch mask. The loss function for the second part is defined as follows:
\begin{equation}L_2=\text{exp} \left\{{|D_{adv}-D_t|\odot(M_O-M_p)}\right\}\end{equation}
where \( M_O \) represents the adversarial scenario mask of the target car. The loss function for computing depth information is calculated as:
\begin{equation}
L_{depth}=L_1+L_2
\end{equation}
The loss function exhibits exponential growth outside the patch boundaries, while within the overlapped patch regions, it demonstrates quadratic growth. The gradient of the exponential function is always greater than that of the quadratic function. With this setup, we ensure that during the backpropagation process, the gradient outside the mask region of the target object is always greater than within the mask region, allowing it to approach the target depth as quickly as possible. For the optimization of the patch content, \( L_{TV} \) serves as a loss function that mitigates abrupt pixel alterations within the generated image, thereby ensuring smoothness across the image. The \(L_{TV}\) term is defined as follows:
\begin{equation}L_{TV}=\sum_{i,j}\sqrt{\left(p_{i+1,j}-p_{i,j}\right)^2+\left(p_{i,j+1}-p_{i,j}\right)^2}\end{equation}
where \( (i, j) \) denotes the pixel coordinates within the patch. Furthermore, \( L_{NPS} \)  represents the non-printable score, given by:
\begin{equation}L_{NPS}=\sum_{i,j}\min_{c\in C}\left(\left\|p_{i,j} - c\right\|_{_2}\right)\end{equation}
where \( (i, j) \) represents the pixel at location \( (i, j) \) in the patch, and \( c \) is the color vector representing the printable color concentration.
Regarding the limitation on patch size, we introduce the area loss \( L_{area} \), which sums up the binary mask to compute the total area of the patch relative to the car patch mask:
\begin{equation}L_{area}=\sum_{i,j}\left({m_p}_{i,j}\right)\end{equation}
We constrain the size of the mask to be consistent with that of a rectangle mask.
The overall loss function for patch optimization is given by:
\begin{equation} \label{adv_loss} L=\lambda_{1}L_{depth}+\lambda_{2}L_{TV}+\lambda_{3}L_{NPS}+\lambda_{4}L_{area}\end{equation}
where, \(\lambda_{1}, \lambda_{2}, \lambda_{3}, \lambda_{4}\) are the hyperparameters we set.

\section{Experiment}

\subsection{Experimental Setup}

\subsubsection{Attack Model and Dataset} 

In the selection of our target models, we refer to models used in previous studies. We opt for three representative self-supervised MDE models: Monodepth2 \cite{godard2019digging}, Depthhints \cite{watson2019self}, and Manydepth \cite{watson2021temporal}. The random scenes used during training are sourced from various outdoor scene images in the KITTI dataset.

\subsubsection{Data Preparation}
\label{data_pre}

We consider real-world factors such as lighting conditions, variations in car size and distance, etc. We apply different brightness levels to the target objects, resize them, add minor rotations, and simulate cars in the real world. For changes in car size and distance, we randomly scale car images within the range of 50\% to 100\% of their original size. Rotation is applied within a range of ±3° to simulate the roll of the car during driving. Random adjustments to the color of car images are limited to a range of 30\% for brightness, 10\% for contrast, and 10\% for saturation.

\subsubsection{Attack Mode}
\label{sec:attack mode}

Due to the flexibility of our target depth setting, our attack strategy can be diversified. 
In addition to the conventional approach of maximizing the distance of the target car, i.e., making the car displace``Further'' than normal state, in this case, we set the depth values of the target area to zero. The second attack we propose is making the target car ``Disappear'', i.e., aiming to make the car's depth same as that of the background. This attack directly sets the random scene depth to the target depth. Additionally, we can bring the car closer to us by setting the depth of the targeted car portion to 1, known as the ``Closer'' attack. Moreover, we can control the distance of the car from us by incrementing or decrementing the car's depth relative to the original depth map. The attack effect is shown in the third and fourth lines of Fig.\ref{fig:one}. In the physical-world, ``Further'' attacks and ``Disappear'' attacks pose more severe threats to autonomous driving systems, making them of greater research significance and value. Furthermore, these attacks are relatively more challenging to implement compared to ``Closer'' attacks.

\begin{figure*}[tbp]
 \begin{minipage}[b]{0.24\textwidth}
		\centering
		\includegraphics[width=\textwidth]{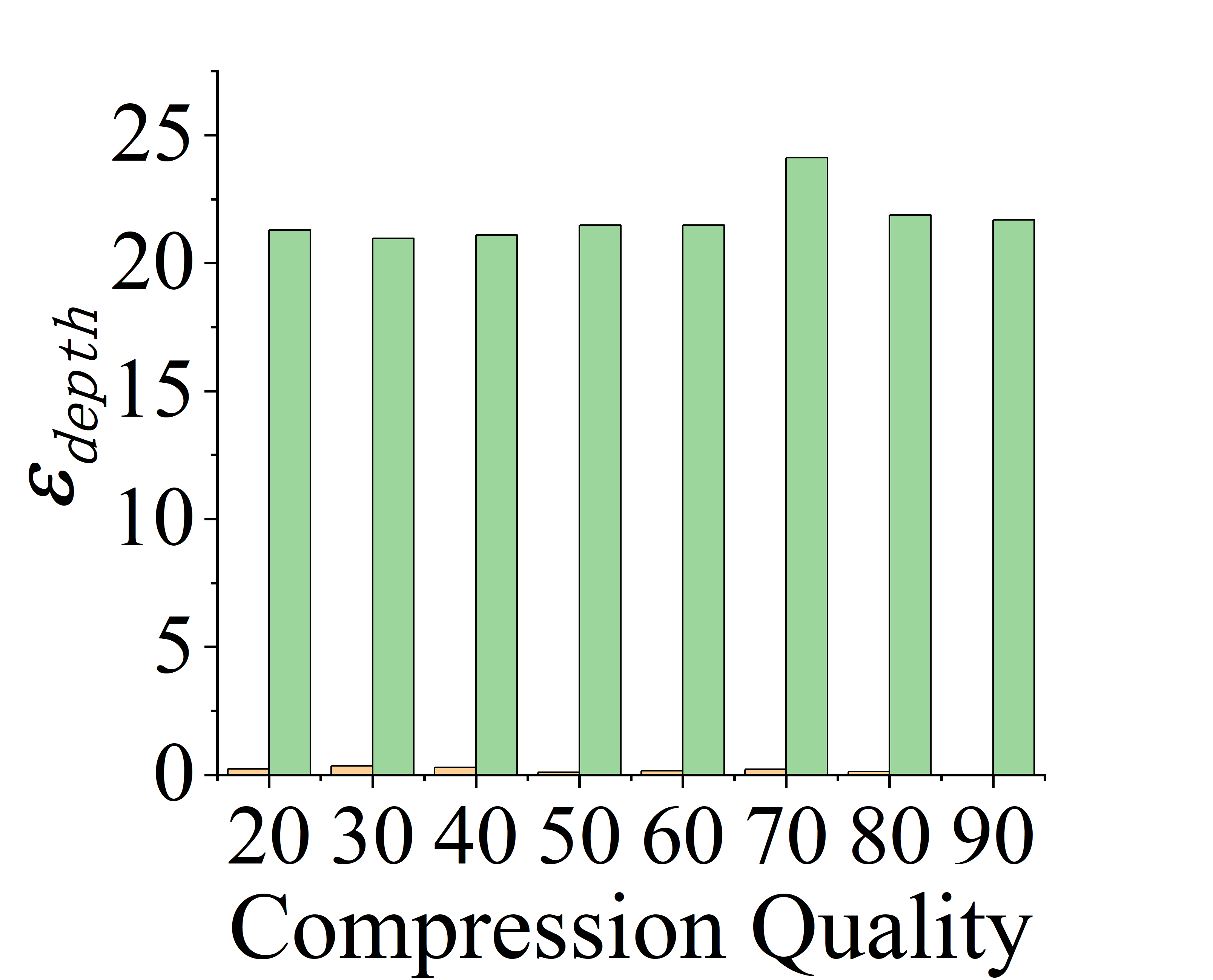}
	\end{minipage}
	\begin{minipage}[b]{0.24\textwidth}
		\centering
		\includegraphics[width=\textwidth]{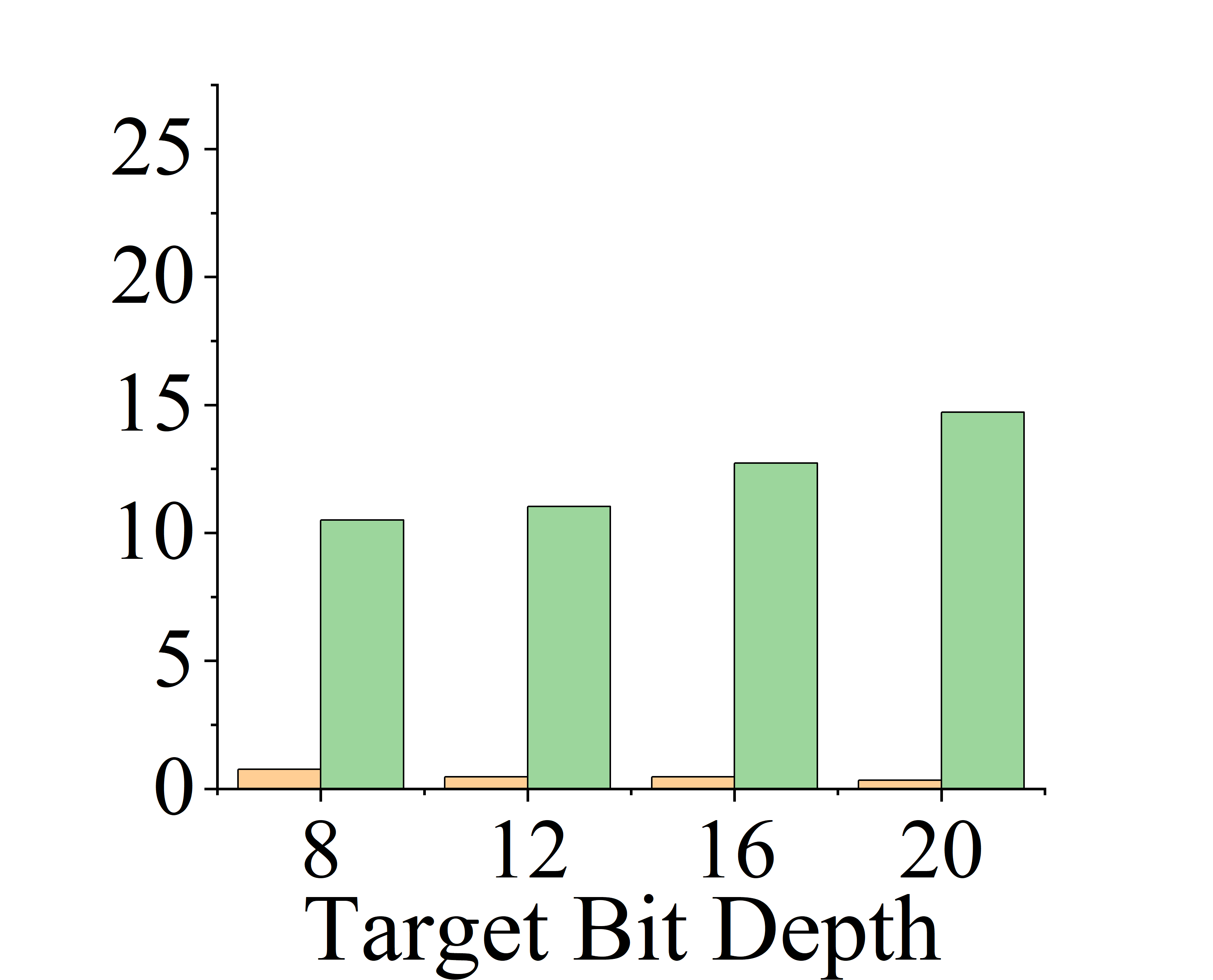}
	\end{minipage}
	\begin{minipage}[b]{0.24\textwidth}
		\centering
		\includegraphics[width=\textwidth]{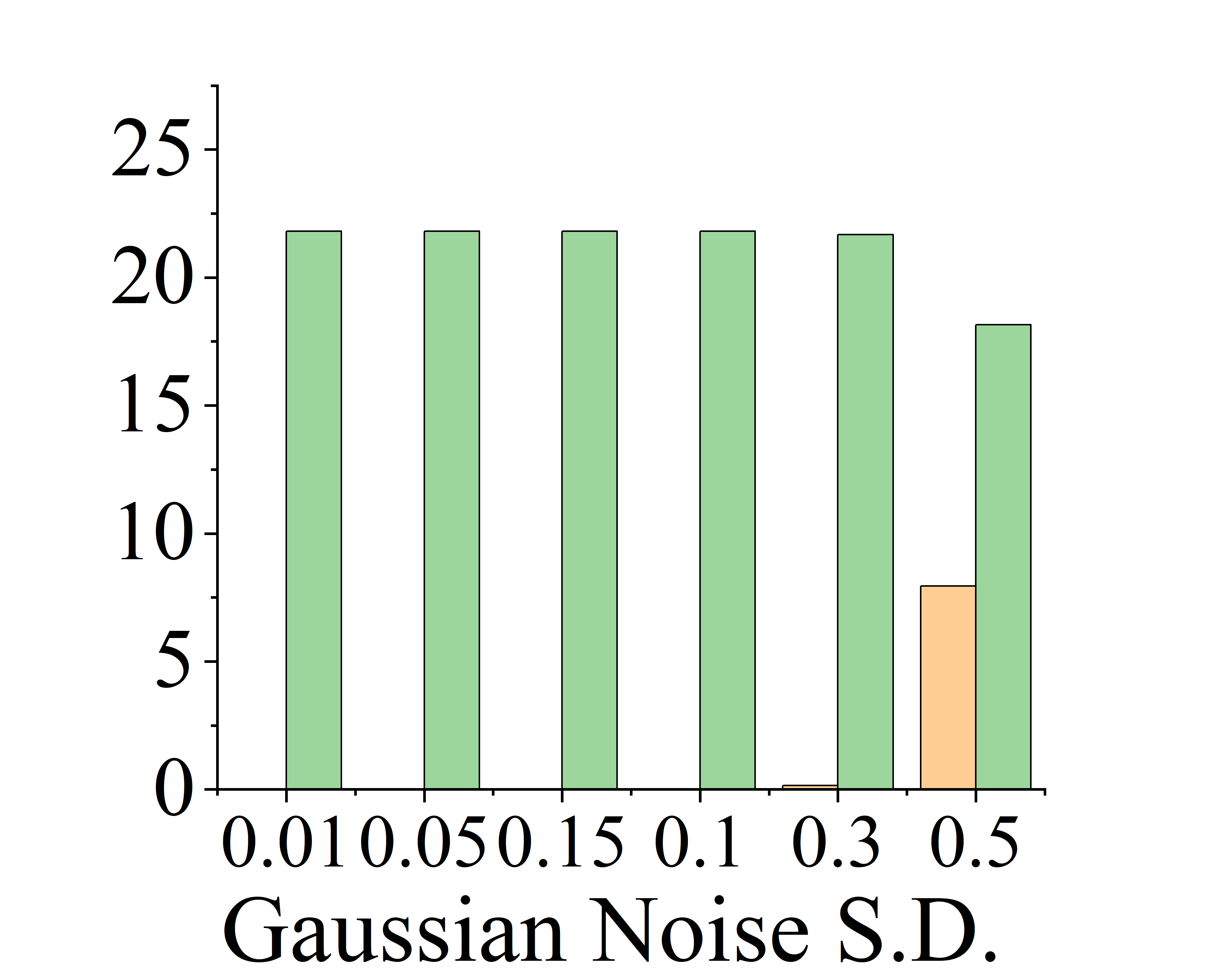}
	\end{minipage}
	\begin{minipage}[b]{0.24\textwidth}
		\centering
		\includegraphics[width=\textwidth]{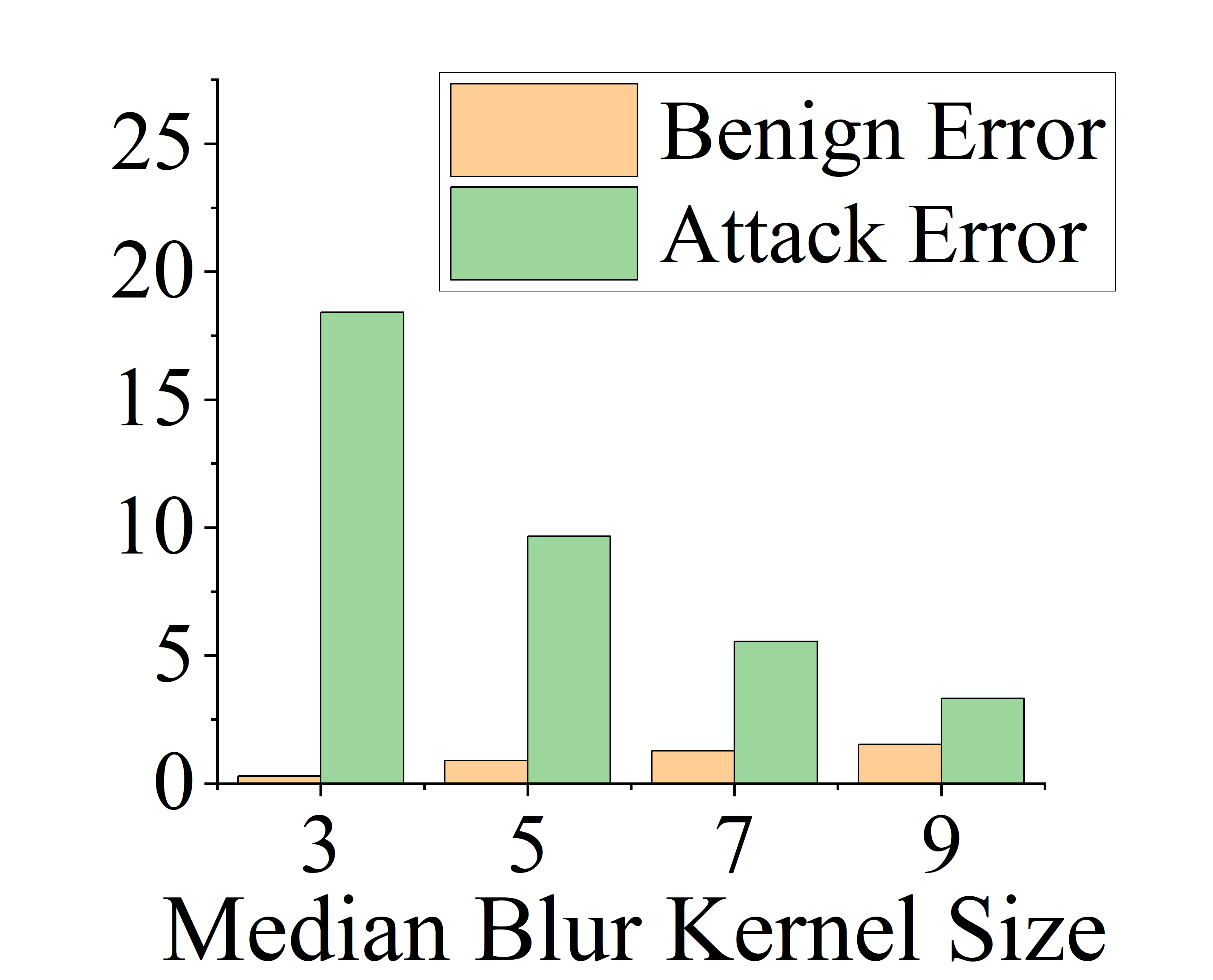}
	\end{minipage}
	\begin{minipage}[b]{0.24\textwidth}
		\centering
		(a)JPEG Compression
	\end{minipage}
	\begin{minipage}[b]{0.24\textwidth}
		\centering
		(b)Bit Depth Reduction
	\end{minipage}
	\begin{minipage}[b]{0.24\textwidth}
		\centering
		(c)Gaussian Noise
	\end{minipage}
	\begin{minipage}[b]{0.24\textwidth}
		\centering
		(d)Median Blur
	\end{minipage}
	\caption{Patch Robustness. Benign Error: Error caused by the defense in benign cases. Attack Error: Error caused by our attack.}
	\label{fig:robustness}
\end{figure*}

\subsubsection{Parameter Settings}

For the optimization of the patch $p$, the four boundaries of the quadrilateral patch mask ${\Theta_1}$, the center coordinates of the circular mask ${\Theta_2}$, as well as the slope $S$ on the boundary of the quadrilateral mask and the radius $R$ of the circular mask, we utilize the Adam optimizer. The learning rate for the patch optimizer is set to 1, with a decay factor of 1/5 applied every fifth optimization step. The learning rates for ${\Theta_1}$, ${\Theta_2}$, $S$, and $R$ are set to 0.01. These learning rates are determined empirically based on multiple experiments. This setting aids in optimizing the masks to form closed contours. We perform a total of 10,000 optimization steps.

\subsubsection{Evaluation Metrics}

We primarily employ five evaluation metrics:
Mean Squared Error(MSE) between the adversarial depth $D_{adv}$ and the target depth $D_t$, where $M_O$ represents the target car mask in the scenario: 
\begin{equation}MSE_t=\frac{\sum\left(D_t-D_{adv}\right)^2 \odot M_{O}}{\sum\left(M_O\right)}\end{equation}
Mean Squared Error (MSE) between the adversarial depth $D_{adv}$ and the depth map of the benign scenario $D_b$:
\begin{equation}MSE_b=\frac{\sum\left(D_b-D_{adv}\right)^2 \odot M_{O}}{\sum\left(M_O\right)}\end{equation}
Proportion of pixels affected in the target object depth region $\alpha$:
\begin{equation}\alpha=\frac{\sum I\left(\left|D_{adv}-D_t\right|\odot M_O\geq0.01\right)}{\sum\left(M_O\right)}\end{equation}
where $I(x)$ is an indicator function, only when x is true, the count is 1. For result validity, we consider pixel depth alteration values exceeding 0.01 as successful attack pixels. Degree of impact on the disparity map of the target region $\varepsilon_{disp}$:
\begin{equation}\varepsilon_{disp}=\frac{{\sum {\left( {\left| {{D_{adv}} - {D_b}} \right|/{D_b}} \right) \odot {M_O}} }}{{\sum {{M_O}} }}\end{equation}
Mean error in the actual depth of the target object region $\varepsilon_{depth}$, $\widetilde{D}_{adv}$ and $\widetilde{D}_b$ represent the real depth of the physical world of adversarial samples and benign samples respectively:
\begin{equation}\varepsilon_{depth}=\frac{{\sum {\left( {\left| {{\widetilde{D}_{adv}} - {\widetilde{D}_b}} \right|/{\widetilde{D}_b}} \right) \odot {M_O}} }}{{\sum {{M_O}} }}\end{equation}
Among the five parameters mentioned above, the first parameter $MSE_t$ represents the proximity to the target depth map, where a smaller value indicates a better attack effect. Conversely, the remaining parameters serve as evaluation metrics, where larger values are indicative of better performance.

\subsection{Main Results of Experiments}

Table \ref{tab:attack model result} compares the effectiveness of our attacks with other models. To facilitate comparison with previous studies, we utilized rectangular masks that can be autonomously optimized in terms of position and shape. The table includes results for three commonly used models in prior research: APARATE~\cite{guesmi2023aparate}, StylePatch~\cite{cheng2022physical}, and the DisM~\cite{yamanaka2020adversarial}. Results for the APARATE model were obtained by replicating the settings from the original paper, while those for StylePatch were derived from running the source code directly. Additionally, results for the DisM were obtained by integrating patches trained by the original authors into our experimental scenes. The best results in the table are highlighted in bold. In our experiments, we attacked the Monodepth2 model and the patch size remained consistent with previous research at 11\% of the image area~\cite{guesmi2023aparate}. Our analysis of the data demonstrates that our model consistently outperforms previous models across all metrics, exhibiting superior effectiveness in patch-based attacks.

Table \ref{tab:at} presents a comparison between different shapes of masks and different attack modes. From the data, it is evident that in the ``Disappear'' attack, the effectiveness of attacks using arbitrary quadrilateral masks has an advantage over rectangular and circular masks. However, in the ``Further'' attack, the effectiveness of attacks using the three shapes of masks varies. Regarding these results, we believe that in the ``Disappear'' attack, the target depth is more complex, and the irregularity in the shape of the quadrilateral mask allows for more diverse depth settings.

\begin{figure}[tbp]
	\centering
	\begin{minipage}[b]{0.25\textwidth}
		\centering
		\includegraphics[width=\textwidth]{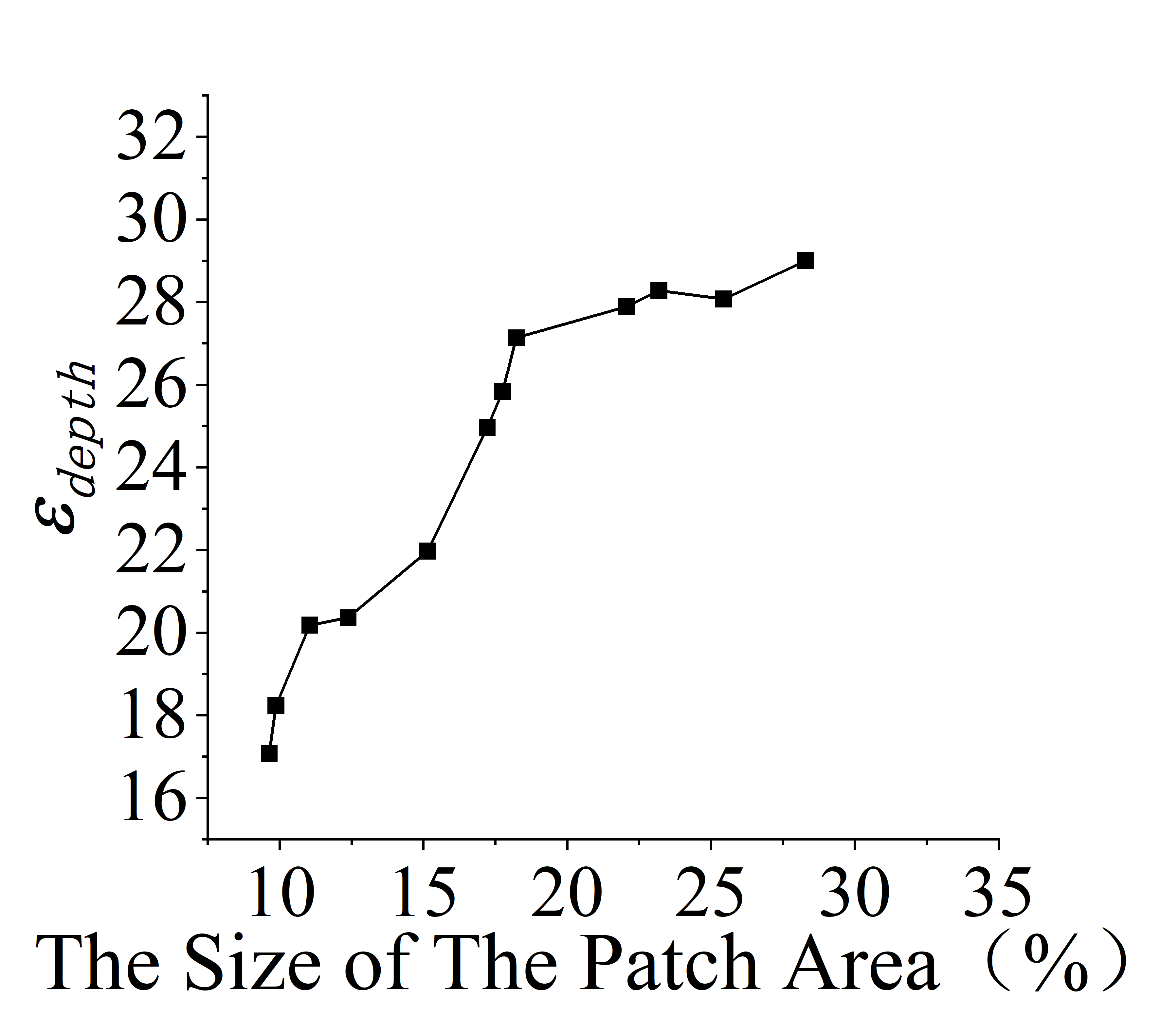}
	\end{minipage}
	\caption{The attack effect changes with the size of the patch area.}
	\label{fig:patch_size}
\end{figure}

We also investigated the impact of patch size on attack effectiveness. In this study, the patches used for the attack were all based on quadrilateral masks and underwent shape and position optimization. The target model for the attack was Monodepth2. As depicted in Fig.\ref{fig:patch_size}, the rate of improvement in patch attack effectiveness gradually slows down with increasing patch area. When the patch area occupies approximately 18\% of the mask area, the patch attack can achieve the best balance between area coverage and attack effectiveness.

\begin{figure*}[t]
	\centering
	\begin{minipage}[b]{0.3\textwidth}
		\centering
		\includegraphics[width=\textwidth]{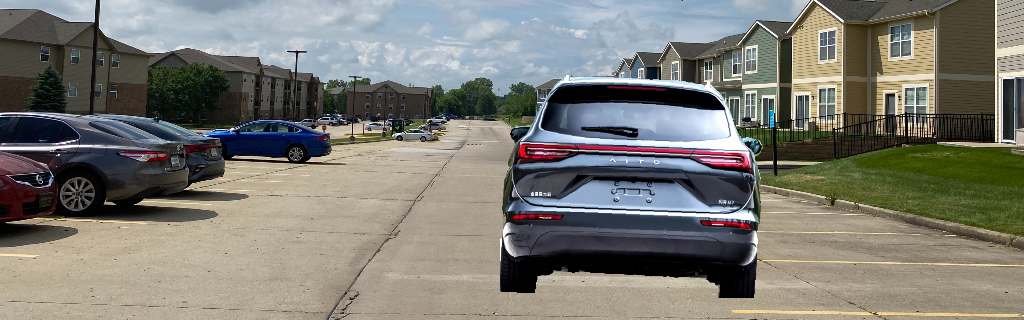}
	\end{minipage}
	\begin{minipage}[b]{0.3\textwidth}
		\centering
		\includegraphics[width=\textwidth]{picture/scar.png}
	\end{minipage}
	\begin{minipage}[b]{0.3\textwidth}
		\centering
		\includegraphics[width=\textwidth]{picture/sdt.png}
	\end{minipage}
	\vspace{0.1cm} % 添加垂直空白
	\begin{minipage}[b]{0.3\textwidth}
		\centering
		\includegraphics[width=\textwidth]{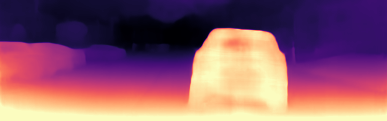}
	\end{minipage}
	\begin{minipage}[b]{0.3\textwidth}
		\centering
		\includegraphics[width=\textwidth]{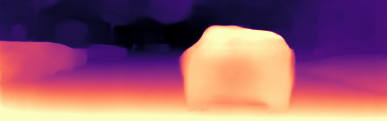}
	\end{minipage}
	\begin{minipage}[b]{0.3\textwidth}
		\centering
		\includegraphics[width=\textwidth]{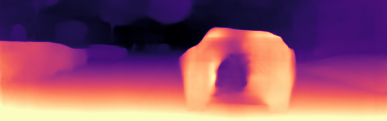}
	\end{minipage}
	\begin{minipage}[b]{0.3\textwidth}
		\centering
		(a)
	\end{minipage}
	\begin{minipage}[b]{0.3\textwidth}
		\centering
		(b)
	\end{minipage}
	\begin{minipage}[b]{0.3\textwidth}
		\centering
		(c)
	\end{minipage}
	\caption{(a) denotes the estimation of benign samples under the initial model , (b) represents the estimation of benign samples under the hardening model, and (c) signifies the estimation of adversarial samples under the hardening model. The hardening model is referenced from Cheng et al.~\cite{cheng2023adversarial}}
	\label{fig:defence}
\end{figure*}

\vspace{-0.2cm}
\subsection{Attack Universality}

\begin{figure}[tbp]
	\centering
	\begin{minipage}[b]{0.22\textwidth}
		\centering
		\includegraphics[width=\textwidth]{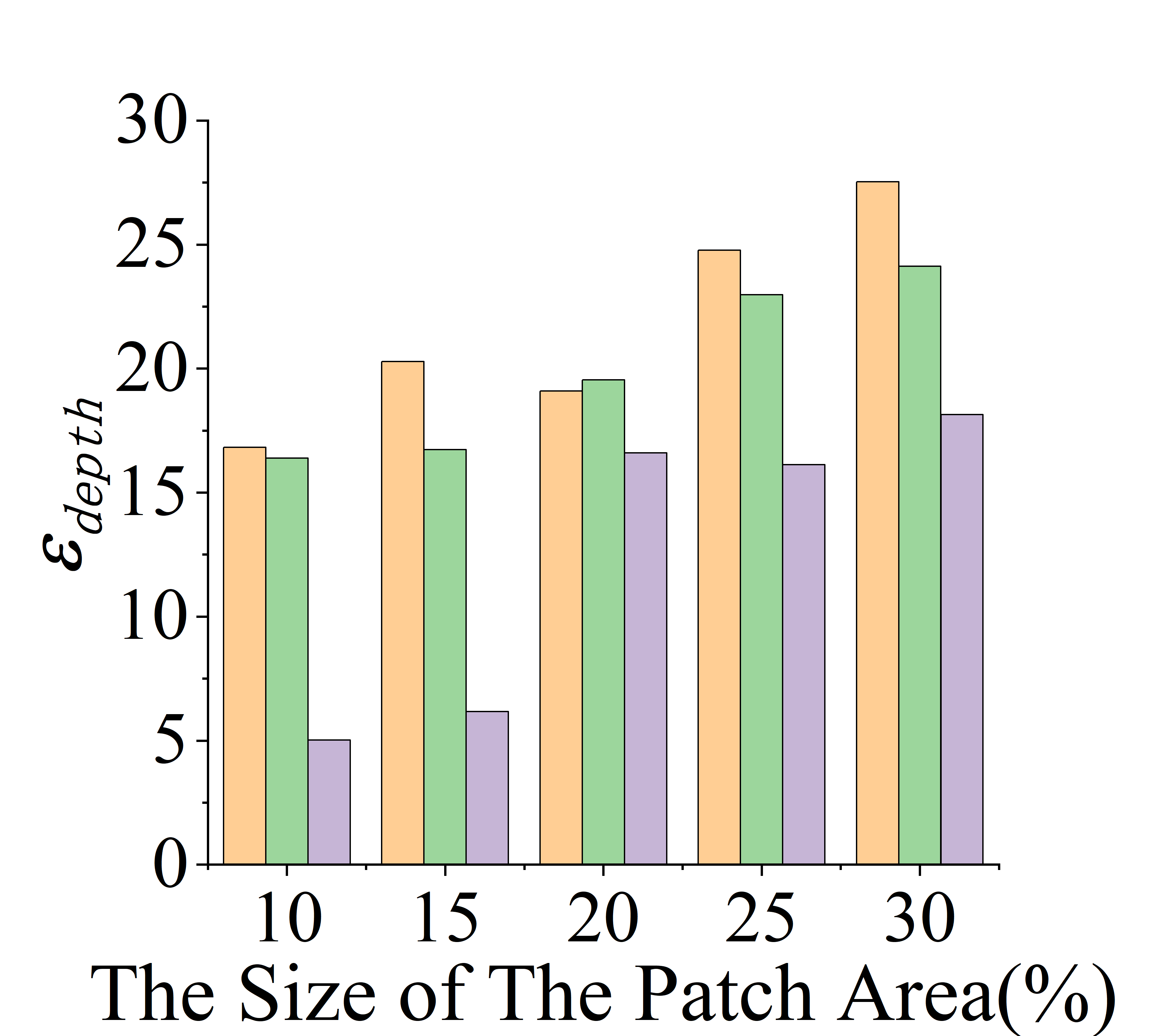}
	\end{minipage}
	\begin{minipage}[b]{0.22\textwidth}
		\centering
		\includegraphics[width=\textwidth]{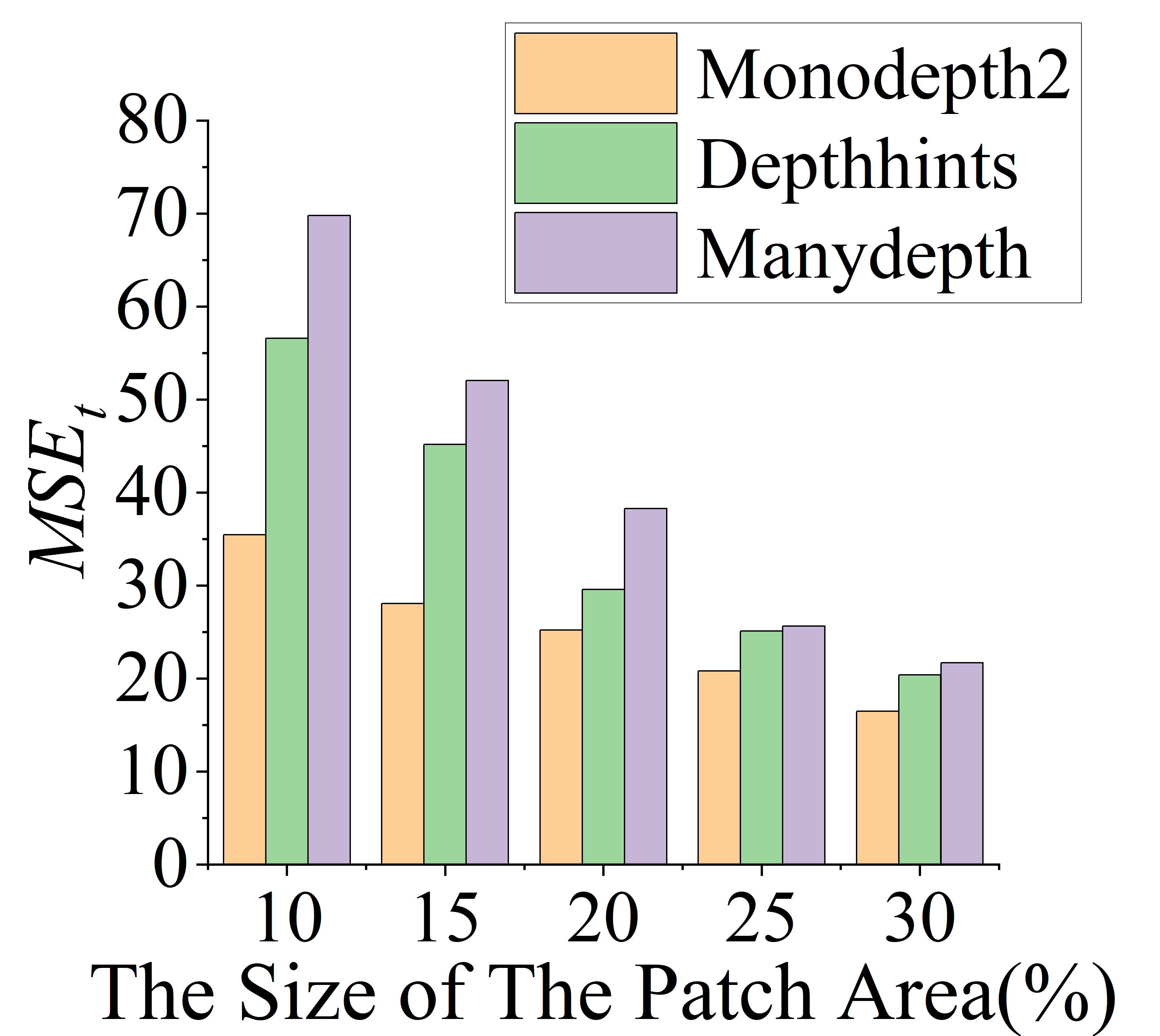}
	\end{minipage}
	\begin{minipage}[b]{0.22\textwidth}
		\centering
		(a)
	\end{minipage}
	\begin{minipage}[b]{0.22\textwidth}
		\centering
		(b)
	\end{minipage}
	\caption{Performance of our attack method on different MDE models. The evaluation index of (a) is the mean error of the true distance from the benign car; (b) Evaluate the performance of our method on different models using $M_{t}$.}
	\label{fig:universality}
\end{figure}

We employed our attack framework to target three pre-selected self-supervised MDE models, testing the generality of our attacks.  Specifically, we conducted attacks using patches of different sizes. We chose $\varepsilon_{depth}$ and the $MSE_{t}$ as the evaluation metrics to validate our experiments and rectangular masks were used. The experimental results, as depicted in Fig.\ref{fig:universality}, confirm the broad efficacy of our attacks.

The empirical findings from our experiments provided compelling evidence regarding the widespread attack efficacy. Across the diverse range of models and evaluation metrics employed, our attacks consistently demonstrated their ability to undermine the integrity and reliability of the targeted self-supervised MDE systems. Such robust performance underscores the universality and potency of our attack framework, highlighting its potential utility in adversarial settings.

\subsection{Robustness}

To assess the robustness of the patches trained in our study, we employed four commonly used adversarial defense methods in the field of deep learning, namely JPEG compression \cite{dziugaite2016study}, bit-depth reduction \cite{xu2017feature}, the addition of Gaussian noise \cite{zhang2019defending}, and median blur \cite{xu2017feature}. The resulting attack outcomes are illustrated in Fig.\ref{fig:robustness}, where the $y$-axis represents the depth error relative to ground truth. For JPEG compression, the $x$-axis denotes the target quality of the compressed image. For bit-depth reduction, the $x$-axis represents the target bit depth. Gaussian noise is characterized by the standard deviation of the added noise, with a mean of 0. The $x$-axis for median blur indicates the size of the median convolution kernel applied.

According to the information presented in the graph, our patches maintain effective attack performance over 10 meters in the majority of cases. However, with the median blur method, although there was a considerable loss in attack effectiveness, notable deviations in depth were also observed in benign samples. This suggests that median blur not only impacts the accuracy of adversarial attacks but also affects the precision of benign samples.

\subsection{Defence Discussion}

The patches we designed exhibit not only stability but also significant attack efficacy against depth estimation defense models. To our knowledge, the only existing work on adversarial defense for monocular depth estimation is by Cheng et al.~\cite{cheng2023adversarial}. We conducted attacks on their fortified monocular depth model, and the experimental results are shown in Fig.~\ref{fig:defence}. Although our attacks are less impactful compared to those on the original model, it is evident that the fortified model exhibits considerable errors when estimating benign samples.

\subsection{Physical Experiment}

\begin{figure*}[ht!]
	\centering
	\begin{minipage}[b]{0.45\textwidth}
		\centering
		\includegraphics[width=\textwidth]{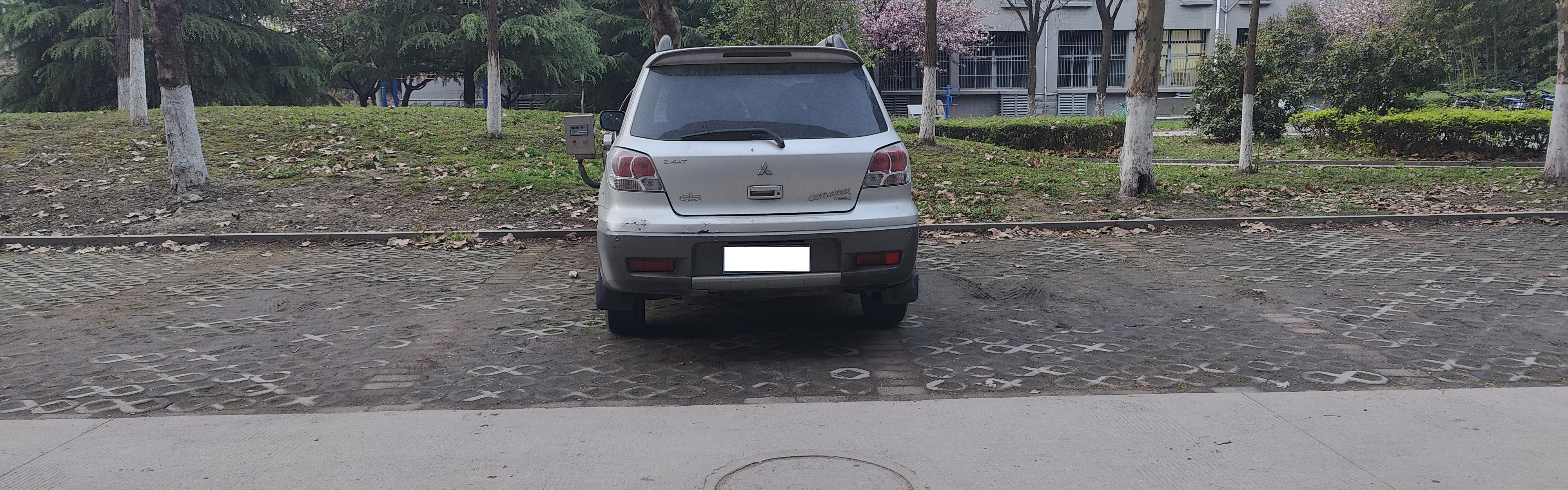}
	\end{minipage}
	\begin{minipage}[b]{0.45\textwidth}
		\centering
		\includegraphics[width=\textwidth]{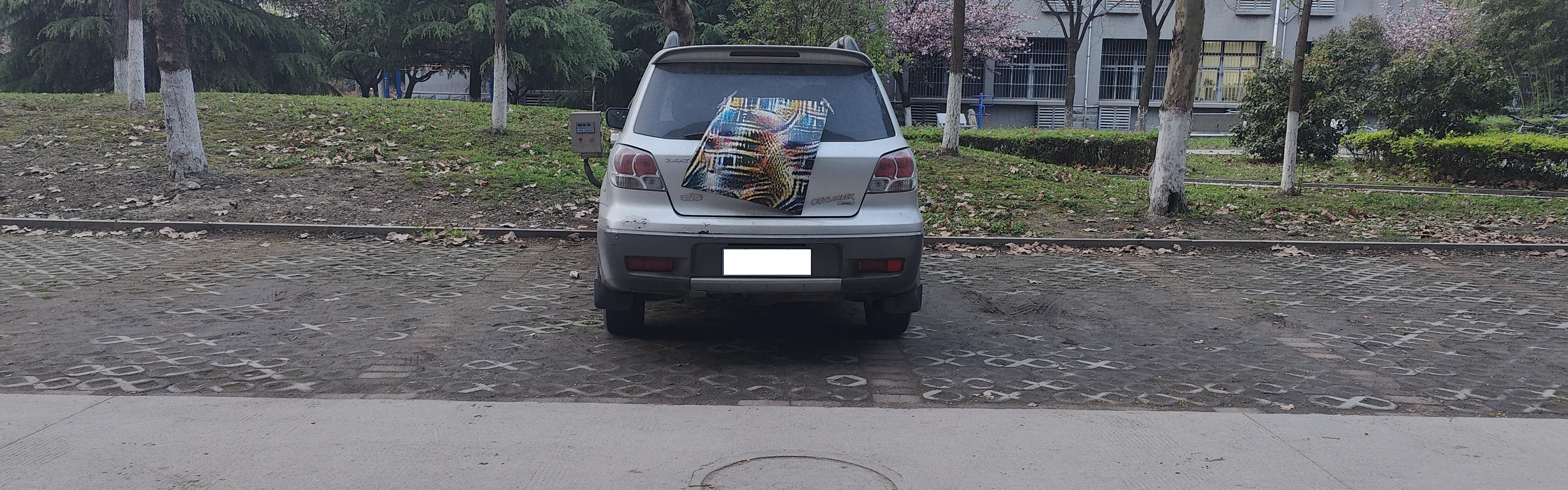}
	\end{minipage}
	\vspace{0.1cm} % 添加垂直空白
	\begin{minipage}[b]{0.45\textwidth}
		\centering
		\includegraphics[width=\textwidth]{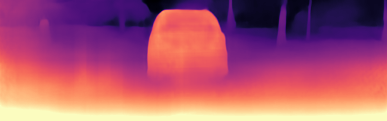}
	\end{minipage}
	\begin{minipage}[b]{0.45\textwidth}
		\centering
		\includegraphics[width=\textwidth]{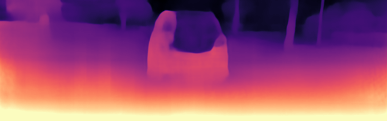}
	\end{minipage}
	\begin{minipage}[b]{0.45\textwidth}
		\centering
		\includegraphics[width=\textwidth]{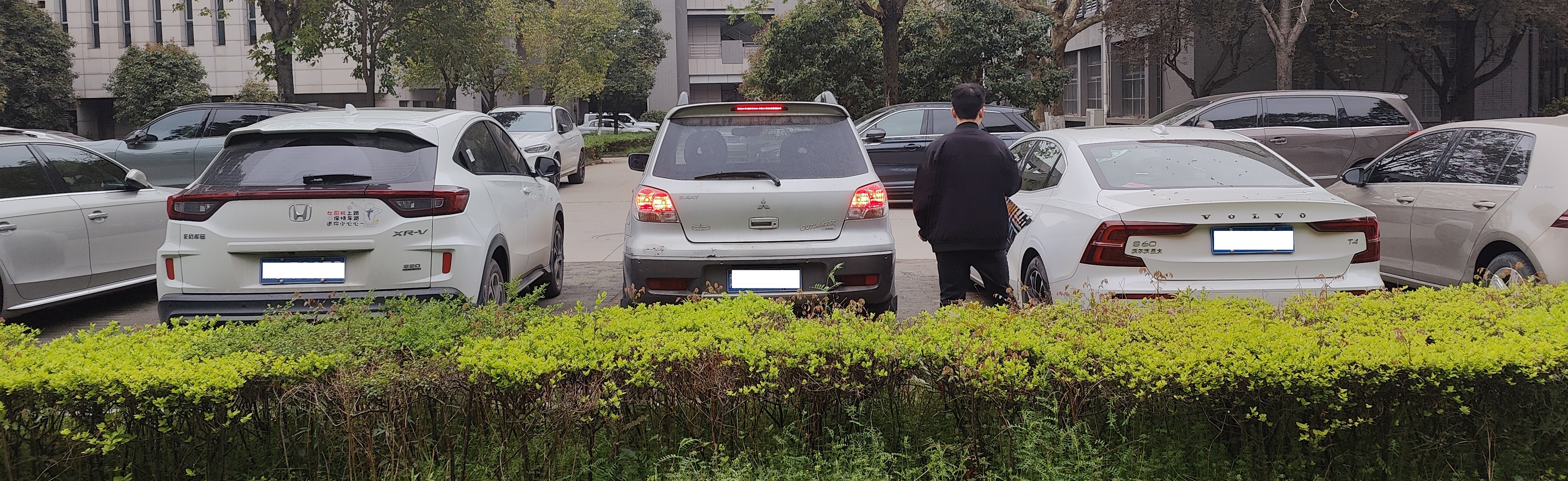}
	\end{minipage}
	\begin{minipage}[b]{0.45\textwidth}
		\centering
		\includegraphics[width=\textwidth]{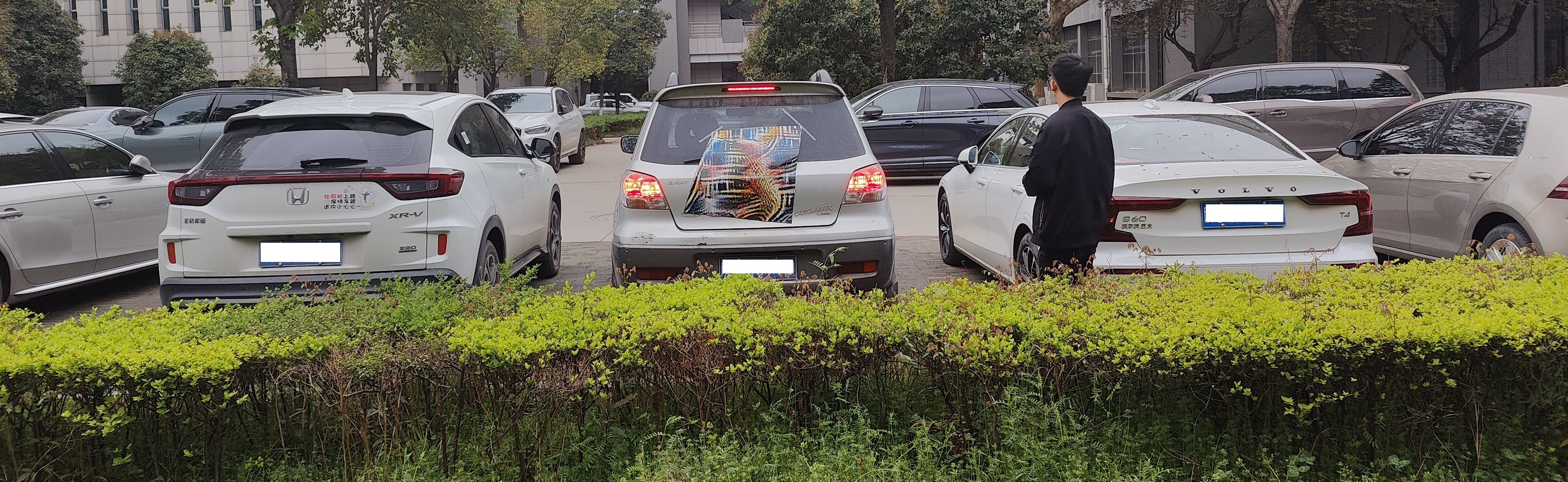}
	\end{minipage}
	\begin{minipage}[b]{0.45\textwidth}
		\centering
		\includegraphics[width=\textwidth]{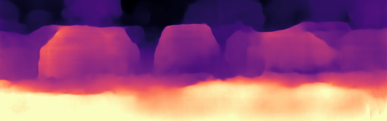}
	\end{minipage}
	\begin{minipage}[b]{0.45\textwidth}
		\centering
		\includegraphics[width=\textwidth]{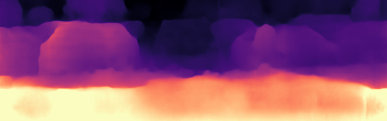}
	\end{minipage}
	\caption{Physical experimental results. The top two rows illustrate the attack effectiveness in single-car scenarios, whereas the bottom two rows represent multi-car attack scenarios. Both types of attacks resulted in an error exceeding 7 meters.}
	\label{fig:physics}
\end{figure*}

In our physical experiments, the target vehicle was systematically positioned at a distance of 7 meters from the camera setup. This separation aligns with the typical braking distance observed at a velocity of 25 miles per hour, a common operational speed encountered within conventional driving environments \cite{vehicle}. The experimental outcomes, graphically depicted in Fig. \ref{fig:physics}, underscore the notable efficacy of our proposed patches across diverse scenarios. Whether it be in single-car scenarios or more intricate scenarios involving multiple vehicles, our patches consistently demonstrated their effectiveness. Notably, both types of simulated attacks yielded errors exceeding 7 meters, further affirming the robustness and reliability of our approach.

\subsection{Ablation Experiment}

The results of our ablation study are presented in Table~\ref{tab:attack model result}. In this study, we first compared the effectiveness of the designed loss functions, testing the attack effectiveness when applying \(L_1\) and \(L_2\) independently. In the table, ``InitRt'' represents the effectiveness of attacks using rectangular masks without position and shape optimization after initialization, while \(L_1\) and \(L_2\) represent the model results when each loss function is individually applied. Additionally, we compared the effectiveness of the two-stage optimization method using differential evolution and the method employing Gaussian kernel convolution for aggregating dispersed pixels. These correspond to the results in the eighth ``DE'' and ninth rows ``Gaussian'' of Table~\ref{tab:attack model result}.

The method of differential evolution was inspired by Wei et al. \cite{10310159}. Firstly, we optimized a patch equal in size to the entire targeted vehicle. Secondly, we froze the content of the patch and selected several points on it. Then, using cubic spline interpolation, these points were connected to form a smooth closed curve. The values inside the curve were set to 1, while those outside were set to 0, forming a patch mask. We utilize the adversarial loss and the patch area loss as the fitness function for differential evolution. We use the differential evolution method to optimize and update the positions of these points, aiming to find the optimal patch shape and position for the attack. In the Gaussian kernel aggregation method, each pixel of the entire mask \( m_p \) is considered as an optimization parameter. Subsequently, gradient-based optimization is initially employed to refine \( m_p \), followed by binarization using \( F_B \) from \ref{equ:fb}. Then, Gaussian kernel convolution is applied to blur the edges, facilitating the aggregation of dispersed pixels. This is succeeded by another binarization step to form a cohesive patch.

Table \ref{tab:attack model result} summarizes the results of various attacks, where the patch size in all experiments constitutes 11\% of the patch mask area. In the DE experiments, we found that optimizing the patch content first and then determining the patch shape significantly reduced the effectiveness of the attack, as the optimization in the first step targeted the entire targeted car, while the patch shape optimized in the second step did not match the patch content. In the Gaussian experiments, we observed difficulty in aggregating noises into a closed graph, indicating challenges in translating digital-space noise attacks into physical-world implementations. Ultimately, consistent across all research findings, our method demonstrates outstanding efficacy in achieving adversarial objectives.
% \vspace{-0.2cm}
% \subsubsection{Limitation}

% Furthermore, in addition to varying the distance of the car, we also attempted to shift the car's position to the left overall, although the results were not satisfactory in achieving the desired attack outcomes.

\section{Limitation}

Our patch has a significant impact on monocular depth estimation networks, enabling alterations in vehicle depth along the front-back axis as designed. However, it is challenging to induce lateral displacements, such as moving a vehicle two meters to the left. Our experiments indicate that patch-based attacks struggle to achieve such effects, leading us to hypothesize that camouflage techniques might yield desired outcomes.

The patches we designed also demonstrate strong attack capabilities in the physical world. However, under intense lighting conditions, the patch content is suppressed by the light, making it difficult for cameras to capture the details completely, thereby reducing the patch's effectiveness. Similarly, in adverse weather conditions such as rain or snow and during periods of low visibility, the camera's inability to fully capture patch details significantly diminishes the attack's impact.

\section{Conclusion}

In this study, we investigate physical-world adversarial patch attacks against MDE. We propose a novel attack framework targeting physical objects, introducing patches that can be optimized in shape and position based on attack effectiveness. Additionally, we can employ different adversarial patches to control the vehicle's distance in the depth map and even cause the depth map of the attacked vehicle to vanish into the background. Experimental results demonstrate the effectiveness and robustness of the patches designed in our study. Furthermore, we explore the impact of patch size on attack effectiveness. Our adversarial patches successfully attack various MDE models and induce multiple types of adversarial depth outputs. We envision that our designed patches will advance the progress of adversarial defense in the field of MDE.

\bibliographystyle{IEEEtran}

\bibliography{jsen}

\end{document}